\documentclass[12pt]{article}

\usepackage[margin=1in]{geometry}
\usepackage[T1]{fontenc}
\usepackage{microtype}
\usepackage{setspace}
\usepackage{etoolbox}

\usepackage{amsmath,amssymb,amsfonts,amsthm,bm}
\usepackage{mathtools}
\usepackage{mathrsfs}
\usepackage{booktabs}
\usepackage{graphicx}
\usepackage{float}
\usepackage{multirow}
\usepackage{array}
\usepackage{enumitem}
\usepackage{placeins}
\usepackage[round,authoryear]{natbib}
\usepackage[hidelinks]{hyperref}
\usepackage{url}
\usepackage[ruled,vlined,linesnumbered]{algorithm2e}

\hypersetup{hypertexnames=false}

\DeclareMathOperator*{\argmin}{arg\,min}
\newtheorem{theorem}{Theorem}
\newtheorem{proposition}{Proposition}
\newtheorem{assumption}{Assumption}
\newtheorem{corollary}{Corollary}
\newcommand{\SMSCP}{\textsc{SMSCP}}

\AfterEndEnvironment{abstract}{%
  \vspace{0.5em}
  \newpage
  \setcounter{page}{1}
  \doublespacing
}

\begin{document}

\singlespacing
\thispagestyle{empty}

\begin{center}
{\LARGE\bfseries State-Matched Sequence Conformal Prediction\par}

\vspace{1.2em}

{\large Ruihang Han$^{a}$ and Li-Hsiang Lin$^{a}$\par}

\vspace{0.5em}

{\large $^{a}$ Department of Mathematics and Statistics, Georgia State University,\\
Atlanta, GA 30303\par}
\end{center}

\vspace{2em}

\begin{abstract}
Sequential conformal prediction needs to decide which past errors should
calibrate uncertainty for the current target. Recency and online error
feedback are useful when recent observations remain relevant, but the
recent past can be a poor reference when comparable states recur at
nonadjacent times, with the recurrence pattern generally unknown.
We propose \emph{State-Matched Sequence Conformal Prediction}
(SMSCP), which constructs an explicit one-sided local-linear state from
pre-target information, uses state similarity to select a matched reference
set of past calibration scores, and then applies the ordinary unweighted
conformal order statistic. In contrast to existing approaches, SMSCP
uses state similarity to identify calibration observations relevant
to the target score quantile. This motivates a theoretical analysis of
whether nearby states have comparable target score quantiles under temporal
dependence. We establish control of the resulting conformal cutoff and
characterize predictive information not captured by the chosen state.
Simulations show that state matching is most effective when score-relevant
states recur across nonadjacent times, while revealing settings in which
its advantage weakens or reverses. Across solar irradiance and bike-demand
benchmarks, SMSCP improves interval score over recency, weighted, adaptive,
and HopCPT comparators, and achieves coverage closer to the nominal $0.90$
level than ResCP on five of six benchmarks.
\end{abstract}

\section{Introduction}

Conformal prediction provides a framework for quantifying predictive
uncertainty with finite-sample guarantees under exchangeability
\citep{vovk2005,shafer2008tutorial}. Standard split conformal and
related sample-splitting methods operate in exchangeable or
independence-type settings
\citep{papadopoulos2002inductive,lei2018,romano2019cqr}, with
cross-fitting and jackknife-style refinements
\citep{barber2021predictive}. Another line localizes or weights
calibration to handle covariate heterogeneity or distribution shift,
including covariate-shift weighting \citep{tibshirani2019conformal},
localized conformal methods \citep{guan2023localized}, and weighted
conformal methods beyond exact exchangeability
\citep{barber2023beyond}. These approaches establish that calibration
relevance need not be uniform across observations.

Sequential conformal prediction raises a different question: which
past observations are relevant for the current target? Early work on
conformal inference under temporal dependence establishes validity
under mixing and related dependence conditions
\citep{chernozhukov2018exact,chernozhukov2021distributional}. Rolling
and online methods such as EnbPI~\citep{xu2021enbpi} and adaptive
conformal inference~\citep{gibbs2021aci,zaffran2022aci} address this
question primarily through temporal recency or online recalibration.
Temporal closeness can be a poor relevance signal, while similar
observed features at different times need not imply similar target score
quantiles. StaLoP \citep{han2026stalop} uses target-local state
similarity to borrow historical panels for mean-function estimation,
whereas SMSCP uses pre-target state similarity to select residual scores
for conformal prediction intervals.

Recent sequential conformal methods move beyond pure temporal recency by
identifying historical observations that are relevant to the current
prediction. HopCPT, ResCP, and RCCP retrieve or reweight past residuals
using similarity in observed or learned representations
\citep{hopcpt2023,neglia2026rescp,jin2026rccp}, while SPCI and KOWCPI
model residual quantiles using residual-history information
\citep{xu2023spci,kowcpi2025}. Such approaches rely on the representation
being sufficiently informative that similar contexts imply similar
residual behavior. This can be restrictive when similar observed
contexts correspond to different residual regimes. SMSCP instead makes
relevance explicitly target-specific by asking whether pre-target states
imply comparable target-score quantiles.

These distinctions motivate \textbf{State-Matched Sequence Conformal
Prediction} (SMSCP). SMSCP makes relevance explicit through a
pre-target state: it constructs an interpretable one-sided local-linear
state summarizing the recent trajectory relative to pre-target inputs,
performs hard Mahalanobis matching, and then applies the ordinary
unweighted conformal order statistic to the matched reference set. The distinction is not
localization itself or the mere use of a state, but where state
similarity enters calibration: it determines reference-set membership
without changing the backbone predictor, score definition, or conformal
quantile rule. The matched reference set is directly inspectable, and
the state coordinates summarize local response level and slopes with
respect to the chosen pre-target inputs.

SMSCP treats conformal calibration as a \emph{state-matching problem}:
select a reference set whose states are comparable to that of the
target, then apply the ordinary conformal rule on that set. The method
is modular: it modifies the calibration reference set, not the backbone
predictor, score definition, or conformal quantile rule. The
local-linear construction is one transparent implementation of this
principle; the theory below treats the resulting pre-target state
abstractly.

Because the matched reference set is target-dependent, classical global
exchangeability does not directly characterize SMSCP. Formally, let
$q(z)$ denote the conditional $(1-\alpha)$ score quantile indexed by
state. Under state--quantile relevance, uniform local concentration over
the target-dependent neighborhoods under temporal dependence, and local
quantile regularity, we bound the cutoff error by a state-matching term,
a sampling-under-dependence term, and a finite-$k$ order-statistic term.
We then compare $q(Z_t)$ with the full-history target quantile through an
explicit state-information discrepancy, which isolates predictive
information not captured by the chosen state. The main theorem therefore
first controls estimation of the state-indexed target $q(Z_t)$;
predictive conditional validity relative to the full pre-target history
additionally requires this discrepancy to vanish. Standard mixing
conditions enter only as sufficient conditions for the required local
concentration, rather than as the form of the main result. The theory does not require
matched scores to be exchangeable or to share their entire conditional
distribution.

Empirically, SMSCP gains are clearest when score-relevant states recur;
they weaken when the state misses relevant information or switching is
abrupt. On real data, SMSCP has lower mean interval score than recency,
weighted, adaptive, and HopCPT comparators; against ResCP, its coverage
estimate is closer to nominal $0.90$ on five of six benchmarks, whereas
ResCP has lower mean interval score on all six. We therefore do not claim
universal dominance.

Our contributions are threefold:
\begin{enumerate}[itemsep=2pt, topsep=2pt, parsep=0pt, partopsep=0pt]
\item We formulate sequential conformal calibration as a
\emph{state-matching problem}, using interpretable pre-target states to
define relevance beyond temporal recency.

\item We propose a modular state-matched conformal algorithm that uses
state similarity only to select a matched reference set and then applies
the ordinary unweighted conformal order statistic without changing the
predictor, score, or conformal quantile rule.

\item We provide quantile-targeted theory and mechanism-isolating
experiments. The theory decomposes cutoff error into a state-matching
term, a sampling-under-dependence term, and a finite-$k$
order-statistic term, while the state-information discrepancy quantifies
predictive information omitted by the chosen state. The experiments use
recency controls together with favorable, mixed, and boundary settings
to probe when state matching helps and when it does not.
\end{enumerate}
\section{Method}
\label{sec:method}

\subsection{Problem setup}
\label{sec:method:setup}

We observe a time-ordered sequence $\{Y_t\}_{t=1}^T$ and construct
prediction intervals for sequential target times at nominal coverage
$1-\alpha$. \SMSCP{} constructs an explicit one-sided local-linear
pre-target state and uses state similarity to identify past calibration
points that are relevant to the current target score quantile. The formal state--quantile relevance condition is stated in
Section~\ref{sec:theory}.

Let $\hat\mu_t$ denote a point predictor of $Y_t$ obtained using only
observations available before time $t$. The nonconformity score is
\begin{equation}
\label{eq:score}
R_t = |Y_t - \hat\mu_t|.
\end{equation}
For a target time $t$, let
\begin{equation}
\label{eq:cal-window}
\mathcal I_{\mathrm{raw}}(t)
=
\{t-n_{\mathrm{cal}},\ldots,t-1\}
\end{equation}
denote the rolling candidate window. The eligible calibration pool
$\mathcal I_{\mathrm{cal}}(t)\subseteq\mathcal I_{\mathrm{raw}}(t)$
contains the indices for which the required pre-target predictor,
state, and any prespecified application-level eligibility condition are
available. Write $n_t=|\mathcal I_{\mathrm{cal}}(t)|$. When all required
quantities are available,
$\mathcal I_{\mathrm{cal}}(t)=\mathcal I_{\mathrm{raw}}(t)$. Since
conformal calibration is performed on the scores $\{R_t\}$, the key
task is to identify past calibration points that are informative for
the current target score quantile.

\subsection{State construction}
\label{sec:method:state}

For each time point $s$, let $X_s\in\mathbb R^p$ be a pre-target input
vector constructed from observations available before $Y_s$, such as
time index, recent lags, covariates, or local summaries of the recent
past. The choice of $X_s$ is application-specific, while \SMSCP{} uses
the same one-sided local-linear state construction below.

For a fixed history length $h$, we define the \SMSCP{} state by
local-linear coefficients \citep{fan1996local}. The coefficients give an
interpretable, one-sided summary of local response level and slopes with
respect to the chosen pre-target inputs. \SMSCP{} uses the resulting
local-linear state as an interpretable representation for calibration
matching. The theory requires that nearby states imply similar target
score quantiles; this state construction is one implementation of that
principle and need not dominate direct matching on the original inputs
in every setting. For analysis, Section~\ref{sec:theory} treats the
resulting state $Z_s$ abstractly. Appendix~\ref{app:state-scale} gives a
sufficient prediction-error scale model.

We construct the state as the one-sided local-linear coefficient vector
\begin{equation}
\label{eq:state-def}
Z_s
=
\argmin_{\boldsymbol\beta\in\mathbb R^{p+1}}
\sum_{r=s-h}^{s-1}
\Biggl[
Y_r-\beta_0
-\sum_{j=1}^p\beta_j(X_{rj}-X_{sj})
\Biggr]^2.
\end{equation}
Here $X_{sj}$ and $X_{rj}$ denote the $j$th coordinates of $X_s$ and
$X_r$. The construction is one-sided: neither $Y_s$ nor future
observations enter the fit. In implementation, the local offsets are
standardized within the history window; this scaling is suppressed in
\eqref{eq:state-def} for notational clarity. The intercept summarizes
the local response level, while the remaining coefficients summarize local
slopes with respect to the pre-target inputs.

\subsection{Conformal prediction based on matched states}
\label{sec:method:srlcp}

At target time $t$, the target state is $Z_t$. We compare each
calibration state $Z_s$, $s\in\mathcal I_{\mathrm{cal}}(t)$, with
$Z_t$ using the Mahalanobis distance
\begin{equation}
\label{eq:distance}
d_{s,t}
=
\sqrt{(Z_s-Z_t)^\top\widehat\Sigma_t^{-1}(Z_s-Z_t)},
\end{equation}
where
\begin{equation}
\label{eq:Sigma}
\widehat\Sigma_t
=
\operatorname{Cov}\!\bigl(
\{Z_s:s\in\mathcal I_{\mathrm{cal}}(t)\}
\bigr)+\varepsilon I,
\qquad \varepsilon>0.
\end{equation}
Here $I$ is the $(p+1)\times(p+1)$ identity matrix. The Mahalanobis
distance accounts for the scale and correlation of the state
coordinates within the eligible calibration pool, and the regularization
term $\varepsilon I$ ensures invertibility.

For a retained fraction $\rho\in(0,1]$, define
\begin{equation}
\label{eq:k}
k_t(\rho)
=
\min\!\left\{
\max\!\left(k_{\min},\lfloor\rho n_t\rfloor\right),
n_t
\right\},
\end{equation}
where $k_{\min}$ is a prespecified lower bound chosen so that the order
statistic below is well defined. We form the matched reference set
\begin{equation}
\label{eq:matched-set}
\mathcal N_t(\rho)
=
\{s\in\mathcal I_{\mathrm{cal}}(t):
\operatorname{rank}(d_{s,t})\le k_t(\rho)\},
\end{equation}
where the rank is computed among
$\{d_{s,t}:s\in\mathcal I_{\mathrm{cal}}(t)\}$, with smaller distance
corresponding to higher state similarity. Ties are broken
deterministically so that $|\mathcal N_t(\rho)|=k_t(\rho)$. The
candidate set $\mathcal R$ for $\rho$ is prespecified. When $\rho=1$,
the procedure reduces to ordinary rolling conformal calibration over
the eligible calibration pool.

The conformal cutoff is
\begin{equation}
\label{eq:quantile}
\hat q_t
=
R_{(m_t)}^{\mathcal N_t(\rho)},
\qquad
m_t=\bigl\lceil(k_t(\rho)+1)(1-\alpha)\bigr\rceil,
\end{equation}
where $R_{(m_t)}^{\mathcal N_t(\rho)}$ denotes the $m_t$th order
statistic of the matched scores
$\{R_s:s\in\mathcal N_t(\rho)\}$. The resulting prediction interval is
\begin{equation}
\label{eq:interval}
\widehat{\mathcal C}_{1-\alpha}(t)
=
[\hat\mu_t-\hat q_t,\;\hat\mu_t+\hat q_t].
\end{equation}
Thus \SMSCP{} matches states, not estimated $q(z)$ values, to form the
reference set and then applies the ordinary unweighted conformal order statistic.

\subsection{Hyperparameter selection}
\label{sec:method:tuning}

We tune the state-construction and matching parameters using the
interval score, a proper scoring rule for prediction intervals
\citep{gneiting2007strictly}. It penalizes interval width together with
the distance of observations falling below or above the interval. For
an interval $[L,U]$ at nominal level $1-\alpha$, define
\begin{equation}
\label{eq:IS}
\begin{aligned}
\mathrm{IS}_\alpha([L,U],Y)
={}&(U-L)\\
&+\frac{2}{\alpha}(L-Y)\mathbf 1\{Y<L\}\\
&+\frac{2}{\alpha}(Y-U)\mathbf 1\{Y>U\}.
\end{aligned}
\end{equation}
We select $(h,\rho)$ by minimizing the mean interval score over a
validation block $\mathcal I_{\mathrm{val}}$, where $\mathcal H$ and
$\mathcal R$ are prespecified finite candidate grids:
\begin{equation}
\label{eq:tuning}
(h^\star,\rho^\star)
=
\argmin_{(h,\rho)\in\mathcal H\times\mathcal R}
\frac{1}{|\mathcal I_{\mathrm{val}}|}
\sum_{t\in\mathcal I_{\mathrm{val}}}
\mathrm{IS}_\alpha\!\bigl(
\widehat{\mathcal C}_{1-\alpha}^{(h,\rho)}(t),Y_t
\bigr).
\end{equation}
The selected pair $(h^\star,\rho^\star)$ is then fixed for final
deployment. The history length $h$ controls how locally the state is
estimated. Smaller $\rho$ retains fewer, closer states, whereas larger
$\rho$ uses more calibration scores while admitting more distant states.
Algorithm~\ref{alg:smscp} in Appendix~\ref{app:algorithm} summarizes
the complete validation-and-deployment procedure.

\section{Theoretical Properties of \SMSCP{}}
\label{sec:theory}

The deployed \SMSCP{} procedure fixes $Z_t$ as the one-sided
local-linear state in Section~\ref{sec:method:state}, but the analysis
treats the resulting state abstractly: nearby states should have similar
target score quantiles. For the observed sequence of state--score pairs
$\{(Z_s,R_s)\}$, the main theorem imposes no specific parametric
time-series model, exchangeability condition, global stationarity of
the observed sequence, or named mixing condition. Temporal dependence
enters only through the uniform local concentration requirement stated
below. The state-information discrepancy introduced later compares the
state-indexed target score quantile with the target based on the full
pre-target history. Accordingly, Theorem~\ref{thm:cutoff} is first a
state-indexed cutoff result; predictive conditional validity additionally
depends on how much target-relevant information the chosen state omits.

The concentration events are understood uniformly over the finite
tuning grid; Appendix~\ref{app:finite-grid} records the corresponding
union-bound argument. For readability, we suppress the candidate index below and write
$k=k_t(\rho)$.

For each evaluation time $t$, let
\[
F_t(r\mid z)=\Pr(R_t\le r\mid Z_t=z)
\]
denote the conditional distribution function of the nonconformity score
$R_t$ given $Z_t=z$, and define its conditional $(1-\alpha)$-quantile by
\[
q_t(z)=\inf\{r:F_t(r\mid z)\ge1-\alpha\}.
\]

\begin{assumption}[State--quantile relevance]
\label{ass:state-quantile}
There exists a common state--quantile function $q$ such that
\[
q_t(z)=q(z)
\]
for all relevant calibration and target times $t$ and states $z$.
Moreover, there exist $r_0>0$, $L_q<\infty$, and $\gamma\in(0,1]$ such
that
\begin{equation}
\label{eq:quantile-holder}
\begin{aligned}
|q(z)-q(z')|
&\le
L_q\|z-z'\|_2^\gamma,\\
&\hspace{-2em}\text{whenever }\|z-z'\|_2\le r_0.
\end{aligned}
\end{equation}
\end{assumption}

Assumption~\ref{ass:state-quantile} is the substantive
state--quantile relevance condition: after conditioning on the
constructed state, the target $(1-\alpha)$ score quantile is stable
across recurrent states and varies smoothly with state distance. It
does not require the response to follow a local-linear model, nor does
it require equality or total-variation closeness of the full conditional
score distributions. Because time and recent history may enter $Z_t$,
the common function $q$ does not require the score quantile to be
constant over calendar time. Appendix~\ref{app:state-scale} gives a
sufficient prediction-error scale model and an explicit failure mode
for this condition.

Let $d_{(k),t}$ denote the $k$th smallest Mahalanobis distance in
\eqref{eq:distance}. Define
\begin{equation}
\label{eq:rstar}
r_t^\star
=
\sqrt{\lambda_{\max}(\widehat\Sigma_t)}\,d_{(k),t}.
\end{equation}
The matched reference set is contained in the Euclidean ball of radius
$r_t^\star$ around $Z_t$. Hence, whenever $r_t^\star\le r_0$,
\[
|q(z)-q(Z_t)|
\le
L_q(r_t^\star)^\gamma
\]
for every state $z$ in that reference set.

For any state neighborhood $A$, define
\[
\widehat F_{t,A}(x)
=
\frac{
\sum_{s\in\mathcal I_{\mathrm{cal}}(t)}
\mathbf 1\{Z_s\in A,\ R_s\le x\}
}{
\sum_{s\in\mathcal I_{\mathrm{cal}}(t)}
\mathbf 1\{Z_s\in A\}
},
\]
and its population counterpart over the same eligible calibration pool by
\[
F_{t,A}(x)
=
\frac{
\sum_{s\in\mathcal I_{\mathrm{cal}}(t)}
\Pr(Z_s\in A,\ R_s\le x)
}{
\sum_{s\in\mathcal I_{\mathrm{cal}}(t)}
\Pr(Z_s\in A)
},
\]
whenever the denominator is positive. Let $\mathcal A$ be the class of
ellipsoids in $\mathbb R^{p+1}$. The target-dependent Mahalanobis
neighborhood used by \SMSCP{} is an element of this fixed,
finite-VC-dimensional class.

\begin{assumption}[Uniform local concentration]
\label{ass:uniform-concentration}
For every fixed $\delta\in(0,1)$, there exists
$\eta_{n_t,k}(\delta)\ge0$ such that, with probability at least
$1-\delta-o(1)$,
\begin{equation}
\label{eq:uniform-concentration}
\begin{aligned}
&\sup_{\substack{A\in\mathcal A:\\
\sum_{s\in\mathcal I_{\mathrm{cal}}(t)}
\mathbf 1\{Z_s\in A\}\ge k}}
\ \sup_{x\in\mathbb R}
\left|
\widehat F_{t,A}(x)-F_{t,A}(x)
\right|\\
&\hspace{8em}\le
\eta_{n_t,k}(\delta).
\end{aligned}
\end{equation}
Along the asymptotic sequence of interest,
$\eta_{n_t,k}(\delta)\to0$.
\end{assumption}

The uniformity in Assumption~\ref{ass:uniform-concentration} is needed
because the target state, Mahalanobis metric, and $k$th-nearest-state
radius are data-dependent. The condition is stated at the empirical-
process level rather than tied to one dependence model.

\begin{assumption}[Local quantile regularity]
\label{ass:quantile-regularity}
There exist $u_0>0$ and $c_q>0$ such that, uniformly over the relevant
evaluation times and states and for all $u\in[0,u_0]$,
\begin{equation}
\label{eq:quantile-regularity}
\begin{aligned}
F_t(q(z)+u\mid z)&\ge1-\alpha+c_q u,\\
F_t(q(z)-u\mid z)&\le1-\alpha-c_q u.
\end{aligned}
\end{equation}
\end{assumption}

Assumption~\ref{ass:quantile-regularity} requires the conditional CDF
to cross the target quantile at a nonvanishing local rate. It is used
only to convert CDF error into quantile error. Appendix~\ref{app:state-scale}
shows that the same prediction-error scale model used to motivate
Assumption~\ref{ass:state-quantile} also implies this condition under a
standard positive-density requirement.

For the theoretical statements below, when $k<n_t$ we assume the $k$th
and $(k+1)$st Mahalanobis distances are distinct; this holds almost
surely under continuous state laws. The deterministic tie rule in
Section~\ref{sec:method:srlcp} remains the implementation rule for
numerical ties.

\begin{theorem}[State-matched cutoff error]
\label{thm:cutoff}
Suppose Assumptions~\ref{ass:state-quantile},
\ref{ass:uniform-concentration}, and
\ref{ass:quantile-regularity} hold. Then, whenever
$r_t^\star\le r_0$ and
\[
\eta_{n_t,k}(\delta)+\frac{2}{k}<\alpha,
\qquad
\frac{1}{c_q}
\left\{
\eta_{n_t,k}(\delta)+\frac{2}{k}
\right\}
\le u_0,
\]
with probability at least $1-\delta-o(1)$,
\begin{equation}
\label{eq:cutoff-bound}
|\hat q_t-q(Z_t)|
\le
L_q(r_t^\star)^\gamma
+
\frac{1}{c_q}
\left\{
\eta_{n_t,k}(\delta)+\frac{2}{k}
\right\}.
\end{equation}
\end{theorem}

The bound separates three sources of error:
\[
\begin{aligned}
&L_q(r_t^\star)^\gamma
&&\text{(state-matching term)},\\
&c_q^{-1}\eta_{n_t,k}(\delta)
&&\text{(sampling-under-dependence term)},\\
&\frac{2}{c_qk}
&&\text{(finite-$k$ order-statistic term)}.
\end{aligned}
\]
Smaller $k$ can make the matched reference set more local, whereas
larger $k$ reduces the finite-$k$ order-statistic term and typically
stabilizes empirical quantile estimation. The additional
small-perturbation conditions in Theorem~\ref{thm:cutoff} are technical
requirements for controlling an empirical order statistic using only
local regularity around the target quantile; they are not restrictions
imposed by the algorithm.

The state need not contain all information in the full pre-target
history. Let $\mathcal F_{t-1}$ denote the information available before
$Y_t$ and define
\[
\begin{aligned}
F_t^R(r)&=\Pr(R_t\le r\mid\mathcal F_{t-1}),\\
q_t^\star&=\inf\{r:F_t^R(r)\ge1-\alpha\},
\end{aligned}
\]
together with the state-information discrepancy
\begin{equation}
\label{eq:state-discrepancy}
\Delta_t
=
|q_t^\star-q(Z_t)|.
\end{equation}
Thus $\Delta_t$ measures predictive information in the full pre-target
history that is not captured by $Z_t$. By Theorem~\ref{thm:cutoff}, on
the same event,
\begin{equation}
\label{eq:full-quantile-bound}
|\hat q_t-q_t^\star|
\le
\Delta_t
+
L_q(r_t^\star)^\gamma
+
\frac{1}{c_q}
\left\{
\eta_{n_t,k}(\delta)+\frac{2}{k}
\right\}.
\end{equation}

Thus Theorem~\ref{thm:cutoff} controls estimation of $q(Z_t)$, whereas
approximation to the full-history target $q_t^\star$ additionally depends
on $\Delta_t$.

\begin{corollary}[Asymptotic predictive conditional coverage]
\label{cor:coverage}
Suppose the conditions of Theorem~\ref{thm:cutoff} hold and, in
addition, $F_t^R$ is continuous at $q_t^\star$, so that
$F_t^R(q_t^\star)=1-\alpha$, and there exists a nondecreasing modulus
$\omega$ such that, uniformly over the relevant evaluation times,
\[
\begin{aligned}
|F_t^R(v)-(1-\alpha)|
&\le
\omega(|v-q_t^\star|),\\
\omega(u)&\to0
\quad\text{as }u\downarrow0.
\end{aligned}
\]
Because $\hat q_t$ is $\mathcal F_{t-1}$-measurable, with probability
at least $1-\delta-o(1)$,
\begin{align}
&\left|
\Pr\!\left\{
Y_t\in\widehat{\mathcal C}_{1-\alpha}(t)
\mid\mathcal F_{t-1}
\right\}
-(1-\alpha)
\right|
\nonumber\\
&\qquad\le
\omega\!\left(
\Delta_t
+
L_q(r_t^\star)^\gamma
+
\frac{1}{c_q}
\left\{
\eta_{n_t,k}(\delta)+\frac{2}{k}
\right\}
\right).
\label{eq:coverage-bound}
\end{align}
Hence predictive conditional coverage converges to the nominal level if
\[
\Delta_t\to_p0,
\qquad
r_t^\star\to_p0,
\qquad
\eta_{n_t,k}(\delta)\to0,
\qquad
k\to\infty.
\]
\end{corollary}

Under stationary absolute regularity ($\beta$-mixing), finite-VC
empirical-process results provide one sufficient condition for
Assumption~\ref{ass:uniform-concentration} under suitable mixing decay
and suitable growth of $k$ \citep{yu1994mixing}; see
Appendix~\ref{app:uniform-concentration}. The appendix also records
sufficient conditions for state--quantile relevance, $\Delta_t=0$, and
shrinking matched neighborhoods.
\section{Experiments}
\label{sec:exp}

\subsection{Setup}
\label{sec:exp:setup}

We evaluate \SMSCP{} on simulation designs and two real-data
benchmarks---solar irradiance and hourly bike demand---at nominal level
$1-\alpha=0.90$. The non-neural simulation panel uses $B=500$ paired
Monte Carlo replications for each of ten core cells formed by five
designs and two chronological splits,
$(100,100,100,100)$ and $(300,300,300,300)$,
ordered as training history, validation, calibration, and test. Three additional simulation designs are evaluated under the large split.

Within the mechanism-controlled non-neural panel, all methods use
the same non-anticipating 20-observation rolling lag-one-correlation
predictor and the same nonconformity score
$R_t=|Y_t-\hat\mu_t|$. At each target, the predictor estimates the
lag-one correlation from the 20 most recent observed responses, clips
the coefficient to $[-0.99,0.99]$, and multiplies the latest response
by that coefficient. The separate HopCPT and ResCP comparisons use the
same frozen point predictions and evaluation targets while retaining
their method-specific calibration mechanisms. Thus differences are not driven
by refitting different point forecasters.

The pre-target inputs used to construct the state are application-specific.
The periodic designs and solar benchmark use chronological trajectory
information, autoregressive and exogenous designs use the corresponding
lags or covariates, and the bike benchmark uses exact previous-day and
previous-week counts. The state-matching and conformal calibration rules
are otherwise unchanged across settings.

The non-neural comparator panel contains \textsc{Rolling}, \textsc{Recent-$k$},
\textsc{Tuned-Recent}, \textsc{Tuned-NexCP}~\citep{barber2023beyond},
\textsc{ACI-tuned}~\citep{gibbs2021aci}, and
\textsc{ECI}~\citep{wu2025eci}.
\textsc{Recent-$k$} is the mechanism control: it uses the same
eligible calibration pool, target-specific reference-set size, predictor,
score, and ordinary conformal order statistic as \SMSCP{}, but selects
by recency rather than by state distance. \textsc{Tuned-Recent}
independently tunes its retained fraction on the same chronological
validation block. \textsc{Tuned-NexCP} independently tunes its decay
parameter. \textsc{ACI-tuned} uses rolling conformal scores and updates
its miscoverage level after each revealed outcome, with its learning
rate selected by validation interval score; \textsc{ECI} independently
tunes its learning rate.
For \SMSCP{}, the implementation searches
$h\in\{5,7,9,11,13\}$ and retained fractions $\rho\in\{0.05,0.08,0.10,0.15,0.20,0.25,0.33,0.40,0.50,1\}$,
with $k_{\min}=10$ and $k_t(\rho)$ defined in~\eqref{eq:k}.
All tuning is performed once on the chronological validation block and
the selected hyperparameters are then fixed throughout test
evaluation. In particular, $\rho=1$ includes the rolling calibration
special case in the \SMSCP{} search space.

HopCPT~\citep{hopcpt2023} is evaluated separately as the
learned-retrieval comparator. Its simulation comparison uses 50
independent data realizations with three fixed neural initializations per
realization, rather than the $B=500$ protocol of the non-neural panel, so
we report it separately instead of mixing estimates from different Monte
Carlo samples in the same table. The three initialization-level results
are averaged within each realization, and paired Monte Carlo inference is
then performed across the 50 independent data realizations. We use 3000
training epochs, matching the published HopCPT experiment configuration.

ResCP~\citep{neglia2026rescp} is evaluated separately on the real-data
benchmarks using the same frozen point predictor, candidate window, and
evaluation targets; implementation and inference details are in
Appendix~\ref{app:rescp-realdata}.

We report empirical coverage, mean interval width, and mean
interval score $\mathrm{IS}_\alpha$ from~\eqref{eq:IS}. For paired
comparisons we define
$\Delta_{\mathrm{IS}}=\mathrm{IS}_{\mathrm{comparator}}-\mathrm{IS}_{\mathrm{SMSCP}}$,
so positive values favor \SMSCP{}. Interval score jointly penalizes interval width and observations outside
the interval, while reporting coverage and mean interval width separately
makes the source of each difference explicit. Complete DGP and tuning details are given in
Appendix~\ref{app:sim_dgp} and Appendix~\ref{app:exp}.

\subsection{Simulation results}
\label{sec:exp:sim}

We focus on the setting \SMSCP{} is designed to exploit: states with
similar target score quantiles recur at nonadjacent times, temporal
proximity is an imperfect relevance signal, and the chosen pre-target
state captures that recurrence. Table~\ref{tab:main_core} summarizes the non-neural
panel.

The clearest gains occur in the recurrent periodic designs. In all four
HomoSin and DualFreq cells, \SMSCP{} has lower mean interval score than
each of the six non-neural comparators, and every paired 95\% Monte
Carlo interval for $\Delta_{\mathrm{IS}}$ is strictly positive.
These gains are not driven by wider intervals: in all four cells,
\SMSCP{} produces narrower mean intervals while maintaining empirical
coverage at or above the nominal level, although it is somewhat
conservative in these recurrent designs. Relative to independently
tuned recency, the interval-score gaps range from $0.217$ to $0.573$.

The remaining designs show where the advantage weakens. HeteroSin-small
is mixed: \textsc{Rolling} has a slightly lower interval score, while
the comparisons with \textsc{Tuned-Recent}, \textsc{Tuned-NexCP},
and \textsc{ACI-tuned} are unresolved; HeteroSin-large favors
\SMSCP{} in interval score but
has slight undercoverage ($0.898$). In ARCH(1), although $Y_{t-1}^2$
contains the conditional-volatility information, the \SMSCP{} state is
constructed by locally regressing the raw response on this input. Because
$\mathbb E(Y_t\mid Y_{t-1}^2)=0$, the resulting level--slope state need
not preserve the volatility scale relevant to the absolute prediction
error. ARCH(1) therefore does not provide clean evidence for the proposed
mechanism. Its small cell is mixed; although \SMSCP{} has lower interval
score than all non-neural comparators in the large cell, this gain is
accompanied by undercoverage ($0.889$) and does not appear consistently
across sample sizes or in the HopCPT comparison.

Abrupt switching gives a different failure mode. In
MarkovSwitch, \textsc{ECI} reduces mean interval score relative to
\SMSCP{} by $0.380$ in the small cell and $0.879$ in the large cell.
\textsc{ACI-tuned} also has lower interval score in the small cell by
$0.040$, but not in the large cell. Thus abrupt regime switching can favor online error-feedback or
learned-retrieval methods, illustrating a setting in which smoothly
recurring state similarity is not the dominant calibration mechanism.

Additional AR and covariate-driven designs in
Appendix~\ref{app:sim_large_extra} show little or no advantage for state
matching when recency or directly observed predictive inputs already
provide an adequate relevance signal.

\begin{figure*}[t]
\centering
\includegraphics[width=0.78\textwidth]{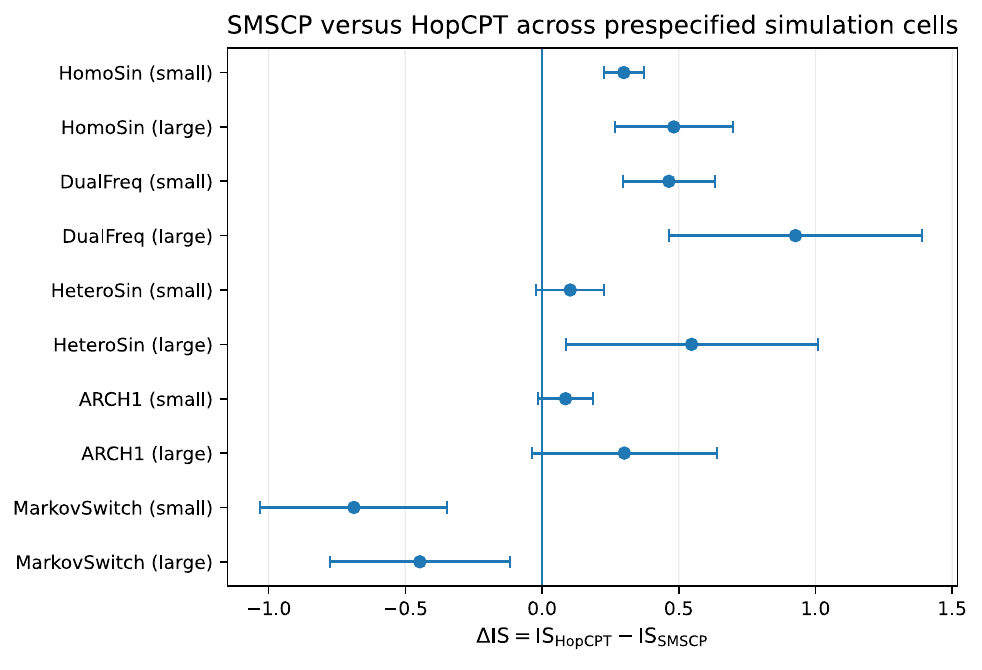}
\caption{Direct comparison with HopCPT across the ten
simulation cells. Points show
$\Delta_{\mathrm{IS}}=\mathrm{IS}_{\mathrm{HopCPT}}
-\mathrm{IS}_{\mathrm{SMSCP}}$ and bars show pointwise 95\% paired
Monte Carlo confidence intervals across 50 independent data
realizations after averaging the three fixed HopCPT
initializations within each realization. Positive values favor
\SMSCP{}.}
\label{fig:hopcpt_sim}
\end{figure*}

Appendix Figure~\ref{fig:hopcpt_sim} gives the direct comparison with the
learned-retrieval comparator. The same pattern appears:
\SMSCP{} is favored in both HomoSin cells, both DualFreq cells, and
HeteroSin-large; HeteroSin-small and both ARCH(1) cells are unresolved;
and HopCPT has lower interval score in both MarkovSwitch cells. In the
four HomoSin/DualFreq cells, \SMSCP{} again produces narrower intervals
and lower interval score than HopCPT while maintaining empirical coverage
at or above the nominal level, although it is somewhat conservative.
Under MarkovSwitch, by contrast, HopCPT obtains lower interval score
through substantially narrower intervals, while its empirical coverage
is farther below the nominal $0.90$ level. The comparison therefore
identifies different settings rather than a uniformly better matching
rule. Taken together, these
numerical patterns are qualitatively consistent with the mechanisms
highlighted by the theory rather than a direct numerical verification of
its decomposition.

\begin{table*}[t]
\centering
\caption{Core non-neural simulation results at nominal $1-\alpha=0.90$.
The small (S) and large (L) splits are $(100,100,100,100)$ and
$(300,300,300,300)$, respectively, with $B=500$ paired Monte Carlo
replications per cell. The first three numeric columns report \SMSCP{}
coverage, mean width, and mean interval score. The remaining columns
report $\Delta_{\mathrm{IS}}
=\mathrm{IS}_{\mathrm{comparator}}-\mathrm{IS}_{\mathrm{SMSCP}}$;
positive values favor \SMSCP{}. Pointwise paired 95\% Monte Carlo
confidence intervals are reported in Appendix~\ref{app:core-full}.}
\label{tab:main_core}
\scriptsize
\setlength{\tabcolsep}{3.4pt}
\resizebox{\textwidth}{!}{%
\begin{tabular}{llrrr|rrrrrr}
\toprule
& & \multicolumn{3}{c|}{\SMSCP{} absolute metrics}
& \multicolumn{6}{c}{$\Delta_{\mathrm{IS}}$ versus \SMSCP{}}\\
\cmidrule(lr){3-5}\cmidrule(lr){6-11}
Design & Size & Cov. & Width & IS
& Rolling & Recent-$k$ & Tuned-Recent & Tuned-NexCP & ACI-tuned & ECI\\
\midrule
HomoSin & S & 0.928 & 2.727 & 2.946 & +0.201 & +0.339 & +0.217 & +0.203 & +0.212 & +0.611 \\
HomoSin & L & 0.933 & 2.625 & 2.817 & +0.315 & +0.414 & +0.317 & +0.317 & +0.319 & +0.527 \\
DualFreq & S & 0.930 & 3.312 & 3.555 & +0.332 & +0.486 & +0.359 & +0.337 & +0.342 & +0.680 \\
DualFreq & L & 0.939 & 3.160 & 3.340 & +0.573 & +0.718 & +0.573 & +0.575 & +0.575 & +0.770 \\
HeteroSin & S & 0.905 & 3.568 & 4.206 & -0.019 & +0.147 & +0.006 & -0.015 & +0.005 & +0.366 \\
HeteroSin & L & 0.898 & 3.385 & 4.044 & +0.129 & +0.183 & +0.131 & +0.131 & +0.136 & +0.434 \\
ARCH(1) & S & 0.894 & 2.997 & 4.126 & -0.045 & +0.090 & +0.044 & -0.046 & -0.039 & +0.358 \\
ARCH(1) & L & 0.889 & 2.852 & 4.127 & +0.036 & +0.154 & +0.076 & +0.035 & +0.028 & +0.143 \\
MarkovSwitch & S & 0.900 & 5.556 & 6.758 & +0.045 & +0.014 & +0.020 & +0.005 & -0.040 & -0.380 \\
MarkovSwitch & L & 0.902 & 5.398 & 6.587 & +0.236 & +0.222 & +0.224 & +0.166 & +0.056 & -0.879 \\
\bottomrule
\end{tabular}}
\end{table*}

\subsection{Real-data benchmark: solar irradiance}
\label{sec:exp:solar}

Solar irradiance tests whether state matching can select calibration
scores more relevant than recency alone: nearby calendar times can have
different local irradiance behavior, whereas distant points with similar
pre-target states may have more comparable target score quantiles.

We evaluate \SMSCP{} on five NSRDB stations~\citep{sengupta2018nsrdb} (Alachua, Phoenix, Seattle,
Houston, and Minneapolis), each providing a full chronological
half-hourly GHI series of length $52{,}608$. The split is
$(n_{\mathrm{train}},n_{\mathrm{val}},n_{\mathrm{cal}},n_{\mathrm{test}})=(2190,2190,2920,45308)$.
Evaluation is restricted to daylight test points defined by solar
zenith angle below $90^\circ$. State construction and point prediction
are performed on the full chronological 30-minute series. For each daylight target, the rolling candidate window is formed first
and then restricted to past daylight points with valid states to obtain
the eligible calibration pool, preventing nighttime zero-irradiance
observations from entering daylight calibration.

Table~\ref{tab:solar_main} shows a stable application-level pattern.
\SMSCP{}
 achieves lower interval score than each of the six non-neural comparators in the panel at all five stations; all 30 pointwise paired moving-block-bootstrap
95\% intervals are strictly positive~\citep{kunsch1989mbb}. Its empirical coverage ranges
from $0.909$ to $0.920$. The gain is not uniformly a width reduction:
the coverage--width pattern differs by station, so coverage and width are
interpreted separately rather than treating higher empirical coverage as
inherently preferable.

The direct learned-retrieval comparison leads to the same station-level
ordering. Across 50 HopCPT initializations on each fixed sequence,
\SMSCP{} has lower mean and median interval score at all five stations.
HopCPT also shows appreciable initialization variability on several
stations. All 50 initializations are included. These summaries describe
initialization variability on the observed sequence, not repeated-sampling
confidence intervals. ResCP results are reported jointly with Bike Sharing
in Appendix~\ref{app:rescp-realdata}.

The original phase-level diagnostic in
Appendix~\ref{app:exp:solar_detail} provides a more granular check.
\SMSCP{} has lower interval score than
\textsc{Rolling}, \textsc{Recent-$k$}, \textsc{Tuned-Recent}, and
\textsc{Tuned-NexCP} in all 20 station--phase combinations, and lower
interval score than \textsc{ECI} in 16 of 20. The five stations are
therefore treated as geographic replication within one solar
application domain, not as evidence of broad multi-domain
generalization.

\begin{table*}[t]
\centering
\caption{Solar irradiance benchmark at nominal $1-\alpha=0.90$.
For the six non-neural comparators,
$\Delta_{\mathrm{IS}}=\mathrm{IS}_{\mathrm{comparator}}
-\mathrm{IS}_{\mathrm{SMSCP}}$ is the paired mean difference on the
fixed station sequence; all 30 corresponding pointwise 7-day
moving-block-bootstrap 95\% intervals are strictly positive. The last two columns
report the median and mean HopCPT-minus-\SMSCP{} interval-score
differences across 50 neural initializations on the same fixed station
sequence; this is initialization variability, not time-series sampling
uncertainty. Width and interval score are in W/m$^2$.}
\label{tab:solar_main}
\scriptsize
\setlength{\tabcolsep}{3.2pt}
\resizebox{\textwidth}{!}{%
\begin{tabular}{lrrr|rrrrrr|rr}
\toprule
& \multicolumn{3}{c|}{\SMSCP{} absolute metrics}
& \multicolumn{6}{c|}{Non-neural $\Delta_{\mathrm{IS}}$}
& \multicolumn{2}{c}{HopCPT $\Delta_{\mathrm{IS}}$}\\
\cmidrule(lr){2-4}\cmidrule(lr){5-10}\cmidrule(lr){11-12}
Station & Cov. & Width & IS
& Rolling & Recent-$k$ & Tuned-Recent & Tuned-NexCP & ACI-tuned & ECI
& Median & Mean\\
\midrule
Alachua & 0.910 & 264.0 & 361.0 & +32.7 & +33.4 & +32.7 & +32.7 & +30.4 & +30.0 & +30.9 & +130.6 \\
Phoenix & 0.920 & 188.2 & 252.8 & +31.5 & +31.5 & +31.5 & +29.9 & +30.4 & +23.4 & +21.1 & +145.8 \\
Seattle & 0.914 & 208.4 & 286.2 & +32.4 & +27.3 & +27.0 & +22.9 & +19.9 & +19.1 & +23.0 & +29.2 \\
Houston & 0.914 & 249.3 & 332.9 & +30.2 & +31.7 & +29.5 & +30.2 & +28.7 & +28.5 & +24.5 & +24.8 \\
Minneapolis & 0.909 & 214.3 & 295.6 & +22.1 & +19.7 & +19.6 & +17.5 & +16.8 & +13.1 & +15.3 & +45.2 \\
\bottomrule
\end{tabular}}
\end{table*}

\subsection{Real-data benchmark: bike sharing}
\label{sec:exp:bike}

The UCI Bike Sharing benchmark~\citep{fanaee2013bike} tests a different form of recurrence:
hourly demand can return to similar daily and weekly conditions even
when the most recent observations are not the most relevant reference
points. We therefore use exact-calendar previous-day and previous-week counts
as strictly pre-target state inputs. No weather variables are used; the
exact state-input definition is given in Appendix~\ref{app:exp:bike_detail}.
Because the source series contains timestamp gaps, the two lags are
matched by timestamp rather than by row offset and are not imputed.
All methods are evaluated on the same 4169 test targets for which the
required daily and weekly lags and resulting state are available. Complete preprocessing, tuning, and inference
details are given in Appendix~\ref{app:exp:bike_detail}.

The resulting coverage, mean interval width, and interval score are reported
in Appendix Table~\ref{tab:app:bike-main}. \SMSCP{} attains coverage $0.894$, mean width $326.4$, and
mean interval score $419.6$. Relative to each of the six non-neural comparators in the panel,
the paired 7-day moving-block-bootstrap interval for
$\Delta_{\mathrm{IS}}$ is strictly positive. The smallest gap among
these comparisons is against \textsc{ECI},
$\Delta_{\mathrm{IS}}=199.5$ with 95\% interval
$[190.1,220.1]$.

HopCPT is evaluated with 50 neural initializations. For the sampling
comparison, its metrics are averaged across initializations within each
target before the same paired moving-block bootstrap is applied.
HopCPT has mean coverage $0.870$, width $317.5$, and interval score
$467.7$, giving
$\Delta_{\mathrm{IS}}=48.2$ (95\% interval $[31.0,63.1]$).
Its mean interval is slightly narrower than \SMSCP{}'s. On the same
target set, \SMSCP{} has lower mean interval score and its empirical
coverage of $0.894$ is closer to the nominal $0.90$ level than HopCPT's
$0.870$.
Against ResCP, the \SMSCP{} coverage point estimate is closer to nominal
$0.90$ on five of six benchmarks (Phoenix excepted), while ResCP has
lower mean interval score on all six. This coverage comparison is
descriptive; full metrics, paired intervals, and Bike 14-day sensitivity
are in Appendix~\ref{app:rescp-realdata}.

\section{Conclusion}

We studied sequential conformal calibration as a target-score-quantile
relevance problem. \SMSCP{} matches past states using an explicit one-sided
local-linear state and then applies the ordinary unweighted conformal order
statistic. The theory separates cutoff error into state matching, dependent
sampling, and finite-$k$ order-statistic terms; full-history predictive
conditional validity additionally requires the state-information discrepancy
$\Delta_t$ to vanish. Experiments are qualitatively consistent with these mechanisms: gains
are strongest when score-relevant states recur, weaken when recency or observed
inputs suffice or the state omits relevant information, and can reverse under
abrupt switching. On these benchmarks, ResCP exposes a coverage--interval-score trade-off. Thus,
\SMSCP{} provides a quantile-specific criterion for calibration relevance
together with an explicit account of its limits.

\begingroup
\singlespacing
\setlength{\bibsep}{1pt}
\bibliographystyle{plainnat}
\bibliography{refs}
\endgroup

\doublespacing
\appendix
\section{Technical Conditions and Algorithmic Details for Section~\ref{sec:theory}}
\label{app:theory}

This appendix summarizes the logic of Theorem~\ref{thm:cutoff} and
Corollary~\ref{cor:coverage}, states sufficient conditions for the main
assumptions, and records the implementation algorithm. Detailed proofs
are omitted from this preprint; the theorem and corollary statements and
all reported numerical results are unchanged.

\subsection{Cutoff bound: main argument}
\label{app:proof-cutoff}

The selected Mahalanobis neighborhood is an element of the ellipsoid
class in Assumption~\ref{ass:uniform-concentration}. State--quantile
relevance bounds the difference between its member quantiles and
$q(Z_t)$; uniform local concentration and quantile regularity transfer
this bound to the empirical order statistic. The finite-$k$ correction
is at most $2/k$, as reflected in Theorem~\ref{thm:cutoff}.

The condition
$\eta_{n_t,k}(\delta)+2/k<\alpha$ is the finite-sample
restriction needed when only local CDF regularity around the
$(1-\alpha)$ quantile is assumed. In particular, very small matched reference sets can place the conformal order statistic at or near the sample
maximum, which cannot be controlled from local quantile behavior alone.

\subsection{Predictive coverage: main argument}
\label{app:proof-coverage}

Because the cutoff is pre-target measurable, predictive conditional
coverage is $F_t^R(\hat q_t)$. Combining the cutoff bound with the
state-information discrepancy $\Delta_t$ and the modulus of continuity
of $F_t^R$ gives Corollary~\ref{cor:coverage}.

\subsection{A prediction-error scale model implying the state assumptions}
\label{app:state-scale}

The main theory is stated directly in terms of the score quantile
because this is the minimal condition needed by \SMSCP{}. The following
result gives a concrete sufficient model linking the constructed state
to that quantile.

\begin{proposition}[State-dependent prediction-error scale]
\label{prop:state-scale}
Define the prediction error generated by the deployed point predictor as
\[
E_t:=Y_t-\hat\mu_t.
\]
Suppose
\begin{equation}
\label{eq:app:scale-model}
E_t=\sigma(Z_t)\xi_t,
\end{equation}
where $Z_t$ and $\hat\mu_t$ are available before $Y_t$,
$\sigma(z)>0$, and the conditional law of $|\xi_t|$ given $Z_t=z$ is a
common distribution $H$ for all relevant $t$ and $z$. Let
\[
c_\alpha=H^{-1}(1-\alpha).
\]
Then
\begin{equation}
\label{eq:app:q-scale}
q_t(z)=q(z)=c_\alpha\sigma(z).
\end{equation}
Consequently, if
\[
|\sigma(z)-\sigma(z')|
\le
L_\sigma\|z-z'\|_2^\gamma,
\]
then Assumption~\ref{ass:state-quantile} holds with
$L_q=c_\alpha L_\sigma$.

If, in addition, $H$ has a density $h$ satisfying
$h(u)\ge h_0>0$ on a neighborhood of $c_\alpha$ and
\[
0<\underline\sigma
\le \sigma(z)\le
\overline\sigma<\infty,
\]
then Assumption~\ref{ass:quantile-regularity} holds on a sufficiently
small uniform neighborhood of the target quantile with
$c_q=h_0/\overline\sigma$.
\end{proposition}

The score distribution under this model obeys
$F_t(r\mid z)=H\{r/\sigma(z)\}$; the quantile and local density
conclusions follow by rescaling. This is a sufficient model, not an
additional requirement for the main theorem.

The prediction errors in Proposition~\ref{prop:state-scale} need not be
serially independent. If the stronger condition
\begin{equation}
\label{eq:app:conditional-magnitude}
\mathcal L(|\xi_t|\mid\mathcal F_{t-1})=H
\end{equation}
holds, then
\[
q_t^\star=c_\alpha\sigma(Z_t)=q(Z_t),
\qquad
\Delta_t=0.
\]
Thus serial correlation is compatible with the theory when it does not
change the conditional magnitude quantile beyond what is captured by
$Z_t$. For example, let $\xi_t=S_tU_t$, where
$S_t\in\{-1,1\}$ is a persistent stationary Markov chain and
$U_t>0$ is i.i.d. and independent of the sign process. Then $\xi_t$ is
serially correlated while $|\xi_t|=U_t$, so
\eqref{eq:app:conditional-magnitude} holds.

Conversely, if the conditional prediction-error scale contains predictable
volatility information not represented by $Z_t$, then $\Delta_t$ need
not vanish. In an ARCH/GARCH-type model, for example, the full-history
quantile is proportional to the current conditional volatility. If the
chosen state does not retain the history needed to determine that
volatility, $q_t^\star$ and $q(Z_t)$ generally differ. This is the
failure mode measured explicitly by \eqref{eq:state-discrepancy}.

\subsection{A standard \texorpdfstring{$\beta$}{beta}-mixing route to uniform local concentration}
\label{app:uniform-concentration}

Assumption~\ref{ass:uniform-concentration} is intentionally stated
without committing the main theorem to a particular time-series model.
Here we record a standard sufficient route based on existing empirical-
process theory.

Write $n=|\mathcal I_{\mathrm{cal}}(t)|$ for the eligible calibration-pool
size and let $W_s=(Z_s,R_s)$. For a measurable set $C$, write
\[
\begin{aligned}
P_n(C)
&=
\frac{1}{n}\sum_{s\in\mathcal I_{\mathrm{cal}}(t)}
\mathbf 1\{W_s\in C\},\\
P(C)
&=\Pr(W_s\in C).
\end{aligned}
\]
where the latter does not depend on $s$ under stationarity. Let
$\mathcal C$ contain the sets
\[
\begin{aligned}
&\{(z,r):z\in A\},\\
&\{(z,r):z\in A,\ r\le x\},
\qquad A\in\mathcal A,\ x\in\mathbb R.
\end{aligned}
\]
Because the state dimension is fixed, $\mathcal C$ has finite VC
dimension.

\begin{proposition}[A $\beta$-mixing sufficient condition]
\label{prop:beta-concentration}
Suppose $\{W_s\}$ is stationary and absolutely regular
($\beta$-mixing) under conditions that yield a uniform VC law of large
numbers for $\mathcal C$, as in \citet{yu1994mixing}. Write
\[
\zeta_n
=
\sup_{C\in\mathcal C}
|P_n(C)-P(C)|.
\]
If $\zeta_n=o_p(k/n)$, then
Assumption~\ref{ass:uniform-concentration} holds with
$\eta_{n,k}=o_p(1)$.
\end{proposition}

Uniform VC convergence controls the numerator and denominator of the
local empirical CDF simultaneously. The lower bound $k/n$ on empirical
neighborhood mass converts the uniform deviation
$\zeta_n=o_p(k/n)$ into the required local concentration.
The dependence-specific VC law is the established result of
\citet{yu1994mixing}.

The proposition is only a sufficient example: the main theorem requires
Assumption~\ref{ass:uniform-concentration} itself and therefore also
covers other dependence structures whenever the same local empirical
concentration can be established.

\subsection{Primitive sufficient condition for a shrinking matched radius}
\label{app:radius}

Uniform local concentration controls score estimation inside a selected
neighborhood. A separate recurrence condition ensures that a
neighborhood containing $k$ calibration states can itself shrink.

\begin{proposition}[Recurrence implies nearest-neighbor shrinkage]
\label{prop:radius}
Let $d=p+1$ be the state dimension. Suppose there exist
$c_->0$ and $r_1>0$ such that, uniformly over relevant target states
$z$ and all $0<r\le r_1$,
\begin{equation}
\label{eq:app:small-ball}
\frac{1}{n}
\sum_{s\in\mathcal I_{\mathrm{cal}}(t)}
\Pr\{\|Z_s-z\|_2\le r\}
\ge
c_-r^d.
\end{equation}
Assume also that the ridge parameter $\varepsilon>0$ in
\eqref{eq:Sigma} is fixed, the uniform empirical-process deviation over
state ellipsoids is $o_p(k/n)$, $k/n\to0$, and
\[
\lambda_{\max}(\widehat\Sigma_t)=O_p(1).
\]
Then
\begin{equation}
\label{eq:app:radius-rate}
\begin{aligned}
d_{(k),t}
&=
O_p\!\left\{(k/n)^{1/d}\right\},\\
r_t^\star
&=
O_p\!\left\{(k/n)^{1/d}\right\}
\to_p0.
\end{aligned}
\end{equation}
\end{proposition}

The small-ball recurrence condition supplies order $k$ historical
states within a Euclidean radius of order $(k/n)^{1/d}$.
The ridge-stabilized metric and concentration condition then imply the
matched-radius rate in \eqref{eq:app:radius-rate}.

The small-ball condition \eqref{eq:app:small-ball} is the formal
recurrence requirement used by the theory: locally around a relevant
current state, the rolling historical pool must retain nonvanishing
probability mass. Proposition~\ref{prop:radius} does not assert that
this condition holds in every nonstationary sequence; it identifies the
setting in which state matching can become increasingly local.
The condition $k/n\to0$ describes one sufficient asymptotic regime,
for example when $\rho_n\to0$ with $n\rho_n\to\infty$; the fixed finite
$\rho$-grids used in the experiments are finite-sample tuning devices
and are not claimed to satisfy this particular regime.

\subsection{Finite-grid validation selection}
\label{app:finite-grid}

No independence between the validation block and later observations is
needed merely to account for selecting $(h,\rho)$ from the finite grid.
Let $M=|\mathcal H\times\mathcal R|$. If a candidate-specific
high-probability statement holds with failure probability at most
$\delta/M+o(1)$ for each candidate, then a union bound gives a
simultaneous event of probability at least $1-\delta-o(1)$. Any
validation-selected pair $(h^\star,\rho^\star)$ therefore inherits the
corresponding bound on that simultaneous event, provided the other
conditions of the relevant theorem are satisfied for the selected
pair. This is the uniformity convention used at the beginning of
Section~\ref{sec:theory}.

\subsection{Algorithmic summary}
\label{app:algorithm}

\begin{algorithm}[H]
\DontPrintSemicolon
\caption{Validation-tuned state-matched sequence conformal prediction}
\label{alg:smscp}
\KwIn{Sequence $\{Y_t\}_{t=1}^T$, nominal level $\alpha$,
rolling candidate-window length $n_{\mathrm{cal}}$, candidate grids
$\mathcal H$ for $h$ and $\mathcal R$ for $\rho$}
\KwOut{Prediction intervals on the test block, or
$\widehat{\mathcal C}_{1-\alpha}(T+1)$}
\BlankLine
Split the sequence into training, validation, calibration, and test blocks\;
Fit a point predictor using only pre-target information and compute
scores $R_t$ by \eqref{eq:score}\;
\ForEach{$h \in \mathcal H$}{
  construct states $\{Z_s^{(h)}\}$ using \eqref{eq:state-def}\;
  \ForEach{$\rho \in \mathcal R$}{
    compute validation intervals by applying
    \eqref{eq:cal-window}, \eqref{eq:matched-set},
    \eqref{eq:quantile}, and \eqref{eq:interval} on
    $\mathcal I_{\mathrm{val}}$\;
    record the mean validation interval score using \eqref{eq:IS}\;
  }
}
select $(h^\star,\rho^\star)$ by \eqref{eq:tuning}\;
\ForEach{target time $t$ in the test block}{
  apply \eqref{eq:cal-window}, \eqref{eq:matched-set},
  \eqref{eq:quantile}, and \eqref{eq:interval} using
  $(h^\star,\rho^\star)$ and output
  $\widehat{\mathcal C}_{1-\alpha}(t)$\;
}
\If{the goal is a single final prediction}{
  apply the same steps to $t=T+1$ and output
  $\widehat{\mathcal C}_{1-\alpha}(T+1)$\;
}
\end{algorithm}


\section{Additional Experimental Results}
\label{app:exp}

The additional results below report the full empirical evaluation.
The simulation results distinguish three settings: state matching helps
when the constructed state captures recurrent score-relevant structure,
adds little when recency already provides relevant calibration scores,
and abrupt switching can favor online error-feedback or learned-retrieval
methods.
For \textsc{ACI-tuned}, the learning rate is selected from
$\{0.0005,0.001,0.002,0.005,0.01,0.02,0.05\}$ by mean validation
interval score, with ties broken toward the smaller value. At test
start $\alpha_t$ is reset to $0.10$, clipped to $[0.001,0.30]$, and
updated after observing the current outcome according to
\[
\alpha_{t+1}
=
\Pi_{[0.001,0.30]}
\left\{
\alpha_t+\eta
\left(0.10-\mathbf 1\{Y_t\notin\widehat{\mathcal C}^{\mathrm{ACI}}_t\}\right)
\right\}.
\]
The real-data sections provide complete station-level solar results and
the common-target Bike Sharing analysis.

\subsection{Complete core simulation results}
\label{app:core-full}
\label{app:width_diagnostics}

Tables~\ref{tab:app:core-small-full} and
\ref{tab:app:core-large-full} report coverage, raw mean interval width,
and mean interval score for every method in the $B=500$
non-neural panel. Table~\ref{tab:app:core-paired} reports the paired
interval-score differences and their pointwise 95\% Monte Carlo
confidence intervals. These tables make the coverage--width source of
each interval-score difference explicit.

\begin{table*}[t]
\centering
\caption{Complete non-neural core simulation results for the small
split $(100,100,100,100)$, $B=500$, nominal $1-\alpha=0.90$.}
\label{tab:app:core-small-full}
\scriptsize
\begin{tabular}{llrrr}
\toprule
Design & Method & Coverage & Mean width & Mean $\mathrm{IS}_\alpha$\\
\midrule
HomoSin & SMSCP & 0.928 & 2.727 & 2.946 \\
 & Rolling & 0.901 & 2.823 & 3.147 \\
 & Recent-k & 0.921 & 3.010 & 3.286 \\
 & Tuned-Recent & 0.903 & 2.843 & 3.163 \\
 & Tuned-NexCP & 0.904 & 2.836 & 3.149 \\
 & ACI-tuned & 0.902 & 2.834 & 3.159 \\
 & ECI & 0.899 & 3.128 & 3.558 \\
\addlinespace
DualFreq & SMSCP & 0.930 & 3.312 & 3.555 \\
 & Rolling & 0.915 & 3.547 & 3.887 \\
 & Recent-k & 0.924 & 3.731 & 4.041 \\
 & Tuned-Recent & 0.917 & 3.576 & 3.915 \\
 & Tuned-NexCP & 0.917 & 3.563 & 3.892 \\
 & ACI-tuned & 0.913 & 3.545 & 3.897 \\
 & ECI & 0.905 & 3.792 & 4.235 \\
\addlinespace
HeteroSin & SMSCP & 0.905 & 3.568 & 4.206 \\
 & Rolling & 0.901 & 3.505 & 4.187 \\
 & Recent-k & 0.912 & 3.736 & 4.353 \\
 & Tuned-Recent & 0.903 & 3.540 & 4.212 \\
 & Tuned-NexCP & 0.903 & 3.528 & 4.191 \\
 & ACI-tuned & 0.901 & 3.535 & 4.211 \\
 & ECI & 0.899 & 3.809 & 4.572 \\
\addlinespace
ARCH(1) & SMSCP & 0.894 & 2.997 & 4.126 \\
 & Rolling & 0.900 & 2.935 & 4.081 \\
 & Recent-k & 0.902 & 3.173 & 4.216 \\
 & Tuned-Recent & 0.900 & 3.068 & 4.170 \\
 & Tuned-NexCP & 0.904 & 2.981 & 4.080 \\
 & ACI-tuned & 0.902 & 3.022 & 4.086 \\
 & ECI & 0.898 & 3.421 & 4.484 \\
\addlinespace
MarkovSwitch & SMSCP & 0.900 & 5.556 & 6.758 \\
 & Rolling & 0.891 & 5.422 & 6.803 \\
 & Recent-k & 0.898 & 5.527 & 6.772 \\
 & Tuned-Recent & 0.898 & 5.518 & 6.778 \\
 & Tuned-NexCP & 0.898 & 5.501 & 6.763 \\
 & ACI-tuned & 0.900 & 5.510 & 6.718 \\
 & ECI & 0.899 & 5.163 & 6.379 \\
\bottomrule
\end{tabular}
\end{table*}

\begin{table*}[t]
\centering
\caption{Complete non-neural core simulation results for the large
split $(300,300,300,300)$, $B=500$, nominal $1-\alpha=0.90$.}
\label{tab:app:core-large-full}
\scriptsize
\begin{tabular}{llrrr}
\toprule
Design & Method & Coverage & Mean width & Mean $\mathrm{IS}_\alpha$\\
\midrule
HomoSin & SMSCP & 0.933 & 2.625 & 2.817 \\
 & Rolling & 0.901 & 2.807 & 3.132 \\
 & Recent-k & 0.920 & 2.966 & 3.231 \\
 & Tuned-Recent & 0.901 & 2.809 & 3.133 \\
 & Tuned-NexCP & 0.902 & 2.815 & 3.134 \\
 & ACI-tuned & 0.902 & 2.813 & 3.135 \\
 & ECI & 0.900 & 2.957 & 3.344 \\
\addlinespace
DualFreq & SMSCP & 0.939 & 3.160 & 3.340 \\
 & Rolling & 0.904 & 3.527 & 3.913 \\
 & Recent-k & 0.925 & 3.762 & 4.059 \\
 & Tuned-Recent & 0.904 & 3.529 & 3.914 \\
 & Tuned-NexCP & 0.905 & 3.536 & 3.916 \\
 & ACI-tuned & 0.904 & 3.530 & 3.916 \\
 & ECI & 0.901 & 3.664 & 4.111 \\
\addlinespace
HeteroSin & SMSCP & 0.898 & 3.385 & 4.044 \\
 & Rolling & 0.900 & 3.474 & 4.173 \\
 & Recent-k & 0.906 & 3.567 & 4.227 \\
 & Tuned-Recent & 0.900 & 3.478 & 4.175 \\
 & Tuned-NexCP & 0.901 & 3.488 & 4.175 \\
 & ACI-tuned & 0.901 & 3.486 & 4.179 \\
 & ECI & 0.899 & 3.717 & 4.478 \\
\addlinespace
ARCH(1) & SMSCP & 0.889 & 2.852 & 4.127 \\
 & Rolling & 0.900 & 2.906 & 4.163 \\
 & Recent-k & 0.902 & 3.105 & 4.281 \\
 & Tuned-Recent & 0.902 & 3.002 & 4.203 \\
 & Tuned-NexCP & 0.903 & 2.951 & 4.162 \\
 & ACI-tuned & 0.900 & 2.994 & 4.155 \\
 & ECI & 0.897 & 3.110 & 4.269 \\
\addlinespace
MarkovSwitch & SMSCP & 0.902 & 5.398 & 6.587 \\
 & Rolling & 0.894 & 5.483 & 6.823 \\
 & Recent-k & 0.894 & 5.474 & 6.808 \\
 & Tuned-Recent & 0.900 & 5.583 & 6.811 \\
 & Tuned-NexCP & 0.902 & 5.550 & 6.752 \\
 & ACI-tuned & 0.900 & 5.440 & 6.643 \\
 & ECI & 0.898 & 4.561 & 5.708 \\
\bottomrule
\end{tabular}
\end{table*}

\begin{table*}[t]
\centering
\caption{Paired interval-score comparisons for the core non-neural
simulation panel. Each cell is
$\Delta_{\mathrm{IS}}=\mathrm{IS}_{\mathrm{comparator}}
-\mathrm{IS}_{\mathrm{SMSCP}}$ followed by its pointwise 95\% paired
Monte Carlo confidence interval. Positive values favor \SMSCP{}.}
\label{tab:app:core-paired}
\scriptsize
\setlength{\tabcolsep}{3pt}
\resizebox{\textwidth}{!}{%
\begin{tabular}{llcccccc}
\toprule
Design & Size & Rolling & Recent-$k$ & Tuned-Recent & Tuned-NexCP & ACI-tuned & ECI\\
\midrule
HomoSin & S & +0.201 [+0.188,+0.213] & +0.339 [+0.325,+0.354] & +0.217 [+0.204,+0.230] & +0.203 [+0.191,+0.215] & +0.212 [+0.200,+0.225] & +0.611 [+0.574,+0.649] \\
HomoSin & L & +0.315 [+0.308,+0.323] & +0.414 [+0.406,+0.422] & +0.317 [+0.309,+0.324] & +0.317 [+0.310,+0.324] & +0.319 [+0.311,+0.326] & +0.527 [+0.519,+0.535] \\
DualFreq & S & +0.332 [+0.315,+0.348] & +0.486 [+0.468,+0.504] & +0.359 [+0.342,+0.377] & +0.337 [+0.320,+0.353] & +0.342 [+0.325,+0.359] & +0.680 [+0.636,+0.724] \\
DualFreq & L & +0.573 [+0.565,+0.581] & +0.718 [+0.710,+0.726] & +0.573 [+0.565,+0.581] & +0.575 [+0.567,+0.584] & +0.575 [+0.567,+0.583] & +0.770 [+0.762,+0.779] \\
HeteroSin & S & -0.019 [-0.034,-0.003] & +0.147 [+0.128,+0.166] & +0.006 [-0.011,+0.023] & -0.015 [-0.031,+0.000] & +0.005 [-0.011,+0.022] & +0.366 [+0.330,+0.402] \\
HeteroSin & L & +0.129 [+0.121,+0.137] & +0.183 [+0.173,+0.192] & +0.131 [+0.124,+0.139] & +0.131 [+0.124,+0.139] & +0.136 [+0.128,+0.143] & +0.434 [+0.423,+0.444] \\
ARCH(1) & S & -0.045 [-0.069,-0.021] & +0.090 [+0.060,+0.120] & +0.044 [+0.014,+0.073] & -0.046 [-0.070,-0.022] & -0.039 [-0.065,-0.014] & +0.358 [+0.293,+0.423] \\
ARCH(1) & L & +0.036 [+0.023,+0.048] & +0.154 [+0.128,+0.181] & +0.076 [+0.061,+0.092] & +0.035 [+0.022,+0.047] & +0.028 [+0.006,+0.050] & +0.143 [+0.125,+0.160] \\
MarkovSwitch & S & +0.045 [+0.010,+0.080] & +0.014 [-0.018,+0.046] & +0.020 [-0.015,+0.054] & +0.005 [-0.028,+0.037] & -0.040 [-0.071,-0.010] & -0.380 [-0.472,-0.287] \\
MarkovSwitch & L & +0.236 [+0.214,+0.258] & +0.222 [+0.200,+0.243] & +0.224 [+0.200,+0.248] & +0.166 [+0.146,+0.185] & +0.056 [+0.037,+0.076] & -0.879 [-0.922,-0.836] \\
\bottomrule
\end{tabular}}
\end{table*}

The four HomoSin/DualFreq cells provide the clearest favorable settings:
all six comparator-minus-\SMSCP{} intervals are strictly positive, and
the raw tables show narrower mean intervals while empirical coverage is
maintained at or above the nominal level, with some conservativeness.
HeteroSin-small is mixed, whereas HeteroSin-large favors \SMSCP{} but has
slight undercoverage. In ARCH(1), $Y_{t-1}^2$ contains
conditional-volatility information, but the raw-response local-linear
state need not preserve the volatility scale relevant to the absolute
prediction error; the large-cell interval-score gain is accompanied by
undercoverage and is not consistent across sample sizes or comparator
classes. MarkovSwitch is the abrupt-switching boundary: ECI has lower
interval score in both sample sizes.

\subsection{Direct learned-retrieval comparison}
\label{app:hopcpt-sim}

Table~\ref{tab:app:hopcpt-sim} reports the full HopCPT comparison,
which tests whether explicit state matching remains competitive with the
learned-retrieval comparator. Each cell uses
50 independent data realizations and three fixed
HopCPT initializations. The three initialization-level results are
averaged within each data realization before paired inference, so the
inferential unit is the data realization rather than the neural seed.

\begin{table*}[t]
\centering
\caption{Direct simulation comparison with HopCPT. The final column is
$\mathrm{IS}_{\mathrm{HopCPT}}-\mathrm{IS}_{\mathrm{SMSCP}}$ with a
pointwise 95\% paired Monte Carlo confidence interval across 50
independent data realizations. The \SMSCP{} metrics in this table are
computed on the same 50 realizations used for the paired HopCPT
comparison and therefore need not equal the $B=500$ averages in
Table~\ref{tab:main_core}.}
\label{tab:app:hopcpt-sim}
\scriptsize
\setlength{\tabcolsep}{2.7pt}
\resizebox{\textwidth}{!}{%
\begin{tabular}{llrrr|rrr|c}
\toprule
& & \multicolumn{3}{c|}{\SMSCP{}} & \multicolumn{3}{c|}{HopCPT}
& $\Delta_{\mathrm{IS}}$ [95\% CI]\\
Design & Size & Cov. & Width & IS & Cov. & Width & IS & \\
\midrule
HomoSin & S & 0.928 & 2.704 & 2.919 & 0.886 & 2.786 & 3.218 & +0.299 [+0.225,+0.372] \\
HomoSin & L & 0.933 & 2.613 & 2.807 & 0.892 & 2.782 & 3.289 & +0.482 [+0.266,+0.699] \\
DualFreq & S & 0.927 & 3.297 & 3.549 & 0.899 & 3.497 & 4.013 & +0.464 [+0.296,+0.632] \\
DualFreq & L & 0.939 & 3.157 & 3.341 & 0.891 & 3.512 & 4.268 & +0.927 [+0.463,+1.391] \\
HeteroSin & S & 0.901 & 3.548 & 4.238 & 0.880 & 3.419 & 4.341 & +0.103 [-0.023,+0.228] \\
HeteroSin & L & 0.893 & 3.342 & 4.060 & 0.875 & 3.389 & 4.607 & +0.547 [+0.087,+1.008] \\
ARCH(1) & S & 0.892 & 3.047 & 4.236 & 0.881 & 2.878 & 4.321 & +0.086 [-0.016,+0.187] \\
ARCH(1) & L & 0.886 & 2.795 & 4.135 & 0.884 & 2.823 & 4.436 & +0.301 [-0.038,+0.640] \\
MarkovSwitch & S & 0.897 & 5.757 & 6.919 & 0.850 & 4.296 & 6.231 & -0.688 [-1.030,-0.346] \\
MarkovSwitch & L & 0.904 & 5.374 & 6.561 & 0.875 & 4.497 & 6.115 & -0.447 [-0.777,-0.116] \\
\bottomrule
\end{tabular}}
\end{table*}

All 1500 HopCPT fits are included. The learned method occasionally
shows substantial initialization variability on particular data
realizations. The comparison supports a narrower statement: explicit
state matching is competitive with the learned-retrieval comparator when
the state captures recurrent relevance, while abrupt switching can favor
that comparator in interval score.

\subsection{Additional simulation designs}
\label{app:robust_sensitivity}
\label{app:sim_large_extra}

Table~\ref{tab:app:robustness} reports the additional large-sample
simulation panel. AR1 is a stationary reference where recency is already
a strong relevance signal. WindowAR tests longer-memory lag structure,
and CovARX tests observed exogenous covariates. Together they probe
settings in which the relevant predictive information is carried by
lags or observed covariates rather than by a recurring time trajectory.

\begin{table*}[t]
\centering
\caption{Additional simulation designs under the large split
$(300,300,300,300)$ with $B=500$ paired replications.}
\label{tab:app:robustness}
\scriptsize
\begin{tabular}{llrrr}
\toprule
Design & Method & Coverage & Mean width & Mean $\mathrm{IS}_\alpha$\\
\midrule
AR1 & SMSCP & 0.899 & 3.524 & 4.460 \\
 & Rolling & 0.900 & 3.520 & 4.447 \\
 & Recent-k & 0.901 & 3.553 & 4.469 \\
 & Tuned-Recent & 0.900 & 3.535 & 4.458 \\
 & Tuned-NexCP & 0.902 & 3.551 & 4.450 \\
 & ACI-tuned & 0.901 & 3.550 & 4.463 \\
 & ECI & 0.898 & 3.618 & 4.546 \\
\addlinespace
WindowAR & SMSCP & 0.892 & 2.063 & 2.644 \\
 & Rolling & 0.900 & 2.104 & 2.645 \\
 & Recent-k & 0.901 & 2.136 & 2.666 \\
 & Tuned-Recent & 0.901 & 2.122 & 2.655 \\
 & Tuned-NexCP & 0.903 & 2.127 & 2.643 \\
 & ACI-tuned & 0.900 & 2.120 & 2.651 \\
 & ECI & 0.898 & 2.184 & 2.733 \\
\addlinespace
CovARX & SMSCP & 0.905 & 1.792 & 2.221 \\
 & Rolling & 0.900 & 1.753 & 2.207 \\
 & Recent-k & 0.902 & 1.776 & 2.225 \\
 & Tuned-Recent & 0.901 & 1.761 & 2.212 \\
 & Tuned-NexCP & 0.903 & 1.766 & 2.208 \\
 & ACI-tuned & 0.901 & 1.758 & 2.213 \\
 & ECI & 0.897 & 1.762 & 2.213 \\
\bottomrule
\end{tabular}
\end{table*}

The stationary AR1 comparison shows little incremental benefit from
state matching: Tuned-Recent is essentially tied with \SMSCP{}, while
Rolling and Tuned-NexCP have slightly lower interval score. WindowAR is
also close to the recency/weighted alternatives. \textsc{ACI-tuned}
is essentially tied with \SMSCP{} in AR1, has slightly higher interval
score in WindowAR, and modestly lower interval score in CovARX. In
CovARX, Rolling, Tuned-Recent, and Tuned-NexCP also have modestly lower
interval score than \SMSCP{}. These results support the intended scope
restriction rather than a universal state-matching advantage.

\subsection{State-space versus feature-space matching}
\label{app:state-feature}

To isolate the effect of the constructed state itself, we compare
State-$k$ with Feature-$k$ while holding fixed the eligible calibration pool,
reference-set size, predictor, score, conformal order statistic, Mahalanobis
metric, and covariance regularization. For WindowAR, Feature-$k$
attains mean interval score $2.6282$ versus $2.6439$ for State-$k$,
with paired Feature-minus-State difference
$-0.0156$ (95\% CI $[-0.0190,-0.0122]$). For CovARX, the corresponding
mean interval scores are $2.2043$ and $2.2205$. Thus the constructed
state is not universally superior to matching on the original
features; it is one interpretable implementation of the matching
principle whose value depends on whether nearby states imply comparable
target score quantiles.

\subsection{Solar benchmark: detailed results}
\label{app:exp:solar_detail}

The five solar stations form one application domain with geographic
replication. Raw NSRDB files~\citep{sengupta2018nsrdb} are converted to full chronological
30-minute series, with reconstructed global horizontal irradiance
$\mathrm{GHI}=\mathrm{DHI}+\mathrm{DNI}\max\{\cos(\mathrm{zenith}),0\}$,
and daylight defined by solar zenith angle below $90^\circ$.
The chronological split uses 2190 observations for the training-history
block, 2190 for validation, 2920 for the rolling candidate-window length, and the
remaining 45308 indices for the test tail. State construction and point
prediction are performed on the full chronological sequence. At each
daylight target, the rolling candidate window is formed first and then
restricted to past daylight points with valid states to obtain the
eligible calibration pool.

The solar state is the same one-sided local-linear level--slope state
used in the time-index simulation designs. With $t_s$ denoting the full
chronological 30-minute index,
$Y_s=\beta_0+\beta_1(t_s-t_t)+\mathrm{error}$ for $s\in[t-h,t-1]$,
with the offset standardized as in the general state construction.
The validation grid uses $h\in\{5,7,9,11,13\}$ and retained fractions
$\rho$ in the grid 0.05, 0.08, 0.10, 0.15, 0.20, 0.25, 0.33,
0.40, 0.50, and 1, with $k_{\min}=10$. Tuned-Recent uses the same $\rho$ grid,
Tuned-NexCP uses decay grid
$\{0.90,0.95,0.97,0.98,0.99,0.995,1\}$, and ECI uses learning-rate
grid $\{0.05,0.10,0.50,1\}$ with burn-in 100. Each hyperparameter is
selected on the chronological validation block and then fixed on test.

\begin{table*}[t]
\centering
\caption{Solar station-level absolute results for the non-neural
panel. Width and interval score are in W/m$^2$.}
\label{tab:app:solar-absolute}
\scriptsize
\begin{tabular}{llrrrr}
\toprule
Station & Method & Coverage & Mean width & Mean $\mathrm{IS}_\alpha$ & $n_{\rm eval}$\\
\midrule
Alachua & SMSCP & 0.910 & 264.0 & 361.0 & 22887 \\
 & Rolling & 0.900 & 266.9 & 393.7 & 22887 \\
 & Recent-k & 0.907 & 282.7 & 394.4 & 22887 \\
 & Tuned-Recent & 0.900 & 266.9 & 393.7 & 22887 \\
 & Tuned-NexCP & 0.900 & 266.9 & 393.7 & 22887 \\
 & ACI-tuned & 0.901 & 272.2 & 391.4 & 22887 \\
 & ECI & 0.897 & 274.5 & 391.0 & 22887 \\
\addlinespace
Phoenix & SMSCP & 0.920 & 188.2 & 252.8 & 22938 \\
 & Rolling & 0.909 & 220.1 & 284.3 & 22938 \\
 & Recent-k & 0.905 & 224.9 & 284.3 & 22938 \\
 & Tuned-Recent & 0.909 & 220.1 & 284.3 & 22938 \\
 & Tuned-NexCP & 0.910 & 225.3 & 282.7 & 22938 \\
 & ACI-tuned & 0.900 & 226.0 & 283.3 & 22938 \\
 & ECI & 0.897 & 220.0 & 276.2 & 22938 \\
\addlinespace
Seattle & SMSCP & 0.914 & 208.4 & 286.2 & 23110 \\
 & Rolling & 0.899 & 209.5 & 318.6 & 23110 \\
 & Recent-k & 0.905 & 213.5 & 313.5 & 23110 \\
 & Tuned-Recent & 0.902 & 207.2 & 313.2 & 23110 \\
 & Tuned-NexCP & 0.910 & 214.2 & 309.0 & 23110 \\
 & ACI-tuned & 0.901 & 217.5 & 306.1 & 23110 \\
 & ECI & 0.896 & 220.7 & 305.3 & 23110 \\
\addlinespace
Houston & SMSCP & 0.914 & 249.3 & 332.9 & 22876 \\
 & Rolling & 0.901 & 245.8 & 363.1 & 22876 \\
 & Recent-k & 0.905 & 266.9 & 364.7 & 22876 \\
 & Tuned-Recent & 0.901 & 249.2 & 362.5 & 22876 \\
 & Tuned-NexCP & 0.901 & 245.8 & 363.1 & 22876 \\
 & ACI-tuned & 0.901 & 251.0 & 361.6 & 22876 \\
 & ECI & 0.896 & 267.5 & 361.5 & 22876 \\
\addlinespace
Minneapolis & SMSCP & 0.909 & 214.3 & 295.6 & 23069 \\
 & Rolling & 0.897 & 212.9 & 317.7 & 23069 \\
 & Recent-k & 0.901 & 215.4 & 315.3 & 23069 \\
 & Tuned-Recent & 0.901 & 213.1 & 315.2 & 23069 \\
 & Tuned-NexCP & 0.906 & 215.1 & 313.1 & 23069 \\
 & ACI-tuned & 0.900 & 222.3 & 312.4 & 23069 \\
 & ECI & 0.897 & 214.3 & 308.7 & 23069 \\
\bottomrule
\end{tabular}
\end{table*}

\begin{table*}[t]
\centering
\caption{Solar paired interval-score differences. Each cell is
comparator-minus-\SMSCP{} mean interval score followed by the 7-day
moving-block-bootstrap pointwise 95\% interval based on 500 bootstrap
replications. All 30 intervals are strictly positive.}
\label{tab:app:solar-mbb}
\scriptsize
\setlength{\tabcolsep}{3pt}
\resizebox{\textwidth}{!}{%
\begin{tabular}{lcccccc}
\toprule
Station & Rolling & Recent-$k$ & Tuned-Recent & Tuned-NexCP & ACI-tuned & ECI\\
\midrule
Alachua & +32.7 [+28.1,+37.2] & +33.4 [+28.8,+38.3] & +32.7 [+28.5,+36.9] & +32.7 [+28.4,+36.8] & +30.4 [+26.0,+35.0] & +30.0 [+25.0,+34.8] \\
Phoenix & +31.5 [+28.7,+34.1] & +31.5 [+28.4,+34.6] & +31.5 [+28.5,+34.4] & +29.9 [+27.2,+32.5] & +30.4 [+27.1,+33.4] & +23.4 [+21.0,+25.7] \\
Seattle & +32.4 [+28.0,+37.8] & +27.3 [+21.6,+33.2] & +27.0 [+22.3,+32.2] & +22.9 [+17.9,+27.8] & +19.9 [+15.1,+24.9] & +19.1 [+14.4,+23.4] \\
Houston & +30.2 [+26.0,+34.8] & +31.7 [+27.5,+36.5] & +29.5 [+25.6,+34.1] & +30.2 [+26.2,+34.6] & +28.7 [+24.9,+32.8] & +28.5 [+24.0,+32.9] \\
Minneapolis & +22.1 [+19.1,+25.0] & +19.7 [+16.2,+23.0] & +19.6 [+16.1,+22.8] & +17.5 [+14.1,+20.7] & +16.8 [+13.5,+20.0] & +13.1 [+9.9,+16.3] \\
\bottomrule
\end{tabular}}
\end{table*}

\begin{table*}[t]
\centering
\caption{Solar comparison with HopCPT across 50 neural initializations
on each fixed station sequence. Mean and median refer to variability
over initialization, not repeated time-series samples.}
\label{tab:app:solar-hopcpt}
\scriptsize
\setlength{\tabcolsep}{3pt}\resizebox{\textwidth}{!}{%
\begin{tabular}{lrrr|rrrrrr}
\toprule
& \multicolumn{3}{c|}{\SMSCP{}}
& \multicolumn{6}{c}{HopCPT across initializations}\\
Station & Cov. & Width & IS
& Mean cov. & Median cov. & Mean width & Median width & Mean IS & Median IS\\
\midrule
Alachua & 0.910 & 264.0 & 361.0 & 0.844 & 0.898 & 244.8 & 261.0 & 491.6 & 391.9 \\
Phoenix & 0.920 & 188.2 & 252.8 & 0.849 & 0.901 & 196.0 & 205.9 & 398.6 & 273.9 \\
Seattle & 0.914 & 208.4 & 286.2 & 0.881 & 0.895 & 206.4 & 205.8 & 315.4 & 309.2 \\
Houston & 0.914 & 249.3 & 332.9 & 0.897 & 0.897 & 239.2 & 239.1 & 357.7 & 357.4 \\
Minneapolis & 0.909 & 214.3 & 295.6 & 0.867 & 0.894 & 200.2 & 206.1 & 340.8 & 310.9 \\
\bottomrule
\end{tabular}}
\end{table*}

A compact phase check gives the same qualitative conclusion without
treating phases as independent datasets. Across the 20 station--phase
combinations, \SMSCP{} has lower interval score than Rolling in 20,
Recent-$k$ in 20, Tuned-Recent in 20, Tuned-NexCP in 20, and ECI in 16.
This heterogeneity is why the main text describes station-specific
coverage--width trade-offs rather than uniform interval narrowing.

\subsection{Bike Sharing benchmark: detailed results}
\label{app:exp:bike_detail}

\begin{table*}[t]
\centering
\caption{Bike Sharing benchmark on the common set of 4169 test targets
at nominal $1-\alpha=0.90$. The first three columns report \SMSCP{}
absolute metrics. The remaining columns report
$\Delta_{\mathrm{IS}}
=\mathrm{IS}_{\mathrm{comparator}}-\mathrm{IS}_{\mathrm{SMSCP}}$.
For the seven comparators listed here, all corresponding
pointwise paired 7-day moving-block-bootstrap 95\% intervals are
strictly positive; full absolute results and intervals are reported in
Appendix~\ref{app:exp:bike_detail}. HopCPT uses the targetwise mean over
50 neural initializations before time-series bootstrap inference.}
\label{tab:app:bike-main}
\scriptsize
\setlength{\tabcolsep}{3.4pt}
\resizebox{\textwidth}{!}{%
\begin{tabular}{rrr|rrrrrrr}
\toprule
\multicolumn{3}{c|}{\SMSCP{} absolute metrics}
& \multicolumn{7}{c}{$\Delta_{\mathrm{IS}}$ versus \SMSCP{}}\\
\cmidrule(lr){1-3}\cmidrule(lr){4-10}
Cov. & Width & IS
& Rolling & Recent-$k$ & Tuned-Recent & Tuned-NexCP & ACI-tuned & ECI & HopCPT\\
\midrule
0.894 & 326.4 & 419.6
& +315.4 & +284.7 & +296.2 & +276.3 & +286.9 & +199.5 & +48.2\\
\bottomrule
\end{tabular}}
\end{table*}


The UCI Bike Sharing hourly series~\citep{fanaee2013bike} contains 17,379 chronologically
ordered observations from 2011--2012, with total rental count
\texttt{cnt} as the response. The source sequence is not perfectly
regular: 75 adjacent timestamps are separated by more than one hour,
with a maximum gap of 37 hours. We preserve the observed sequence,
without inserting or deleting rows. The point predictor uses preceding
observed responses, while state lags are defined by exact elapsed
calendar time.

The chronological split uses 2160 observations for training history,
2160 for validation, 8760 for the calibration block, and the remaining
tail for test evaluation. The analysis uses only the timestamp and
rental count; weather and contemporaneous calendar covariates are not
used. The state inputs are
$X_t=(Y_{t-24\mathrm{h}},Y_{t-168\mathrm{h}})$,
where both entries must be available at the exact corresponding
timestamps. Missing daily or weekly lags are not imputed. For a history
length $h$, the three-dimensional state
$Z_t=(\beta_0,\beta_1,\beta_2)$ is obtained from
\[
\begin{aligned}
Y_s
&=
\beta_0
+
\beta_1
\frac{Y_{s-24\mathrm{h}}-Y_{t-24\mathrm{h}}}{s_1}\\
&\quad+
\beta_2
\frac{Y_{s-168\mathrm{h}}-Y_{t-168\mathrm{h}}}{s_2}
+
\mathrm{error},\\
&\hspace{2em}s\in[t-h,t-1].
\end{aligned}
\]
where $s_1$ and $s_2$ are the corresponding local empirical standard
deviations with floor $10^{-4}$. This state represents daily and weekly
recurrence using only information available before the target.

The \SMSCP{} validation grid uses $h\in\{5,7,9,11,13\}$ and retained fractions
$\rho$ in the grid 0.05, 0.08, 0.10, 0.15, 0.20, 0.25, 0.33,
0.40, 0.50, and 1, with $k_{\min}=10$. Validation selects
$h^\star=7$ and $\rho^\star=0.05$.
Tuned-Recent uses the same $\rho$ grid, Tuned-NexCP uses decay grid
$\{0.90,0.95,0.97,0.98,0.99,0.995,1\}$, and ECI uses learning-rate
grid $\{0.05,0.10,0.50,1\}$ with burn-in 100. For \textsc{ACI-tuned},
the update is run sequentially on the original 4299 valid test targets
before restriction to the final 4169 common targets. HopCPT uses 50
neural initializations and 3000 training epochs.

Exact-lag availability yields 4169 valid \SMSCP{} test targets. All
methods are restricted to this same target set for comparison. Table~\ref{tab:app:bike-absolute} reports absolute metrics.

\begin{table*}[t]
\centering
\caption{Bike Sharing benchmark: absolute results on the common set of
4169 test targets. HopCPT metrics are averaged across 50 neural
initializations within each target.}
\label{tab:app:bike-absolute}
\scriptsize
\begin{tabular}{lrrr}
\toprule
Method & Coverage & Mean width & Mean $\mathrm{IS}_\alpha$\\
\midrule
SMSCP & 0.894 & 326.4 & 419.6\\
Rolling & 0.868 & 397.3 & 734.9\\
Recent-$k$ & 0.907 & 502.4 & 704.3\\
Tuned-Recent & 0.899 & 494.3 & 715.8\\
Tuned-NexCP & 0.912 & 507.4 & 695.9\\
ACI-tuned & 0.906 & 501.9 & 706.4\\
ECI & 0.903 & 433.6 & 619.1\\
HopCPT & 0.870 & 317.5 & 467.7\\
\bottomrule
\end{tabular}
\end{table*}

For time-series sampling inference, HopCPT metrics are first averaged
across the 50 initializations within each target. We then apply the same
paired moving-block bootstrap~\citep{kunsch1989mbb} to all method comparisons, using
$B=500$ replications and blocks spanning seven calendar days. Because
the observed series contains timestamp gaps, the number of evaluated
targets per seven-day block varies; the 4026 admissible starting blocks
contain 101--168 common targets, with median 168.
Table~\ref{tab:app:bike-mbb} reports the resulting pointwise 95\%
intervals.

\begin{table*}[t]
\centering
\caption{Bike Sharing paired 7-day moving-block-bootstrap comparisons,
$B=500$. Differences are comparator-minus-\SMSCP{} on the same 4169
targets. Positive interval-score differences favor \SMSCP{}.}
\label{tab:app:bike-mbb}
\scriptsize
\setlength{\tabcolsep}{3pt}
\resizebox{\textwidth}{!}{%
\begin{tabular}{lccc}
\toprule
Comparator
& $\Delta$Coverage [95\% CI]
& $\Delta$Width [95\% CI]
& $\Delta_{\mathrm{IS}}$ [95\% CI]\\
\midrule
Rolling
& -0.0261 [-0.0368,-0.0175]
& +70.8 [+44.3,+95.7]
& +315.4 [+302.3,+344.7]\\
Recent-$k$
& +0.0127 [+0.0024,+0.0227]
& +176.0 [+163.8,+190.4]
& +284.7 [+274.6,+306.4]\\
Tuned-Recent
& +0.0048 [-0.0038,+0.0133]
& +167.9 [+145.2,+190.1]
& +296.2 [+285.8,+312.8]\\
Tuned-NexCP
& +0.0178 [+0.0089,+0.0297]
& +181.0 [+168.2,+203.3]
& +276.3 [+262.4,+304.5]\\
ACI-tuned
& +0.0120 [+0.0029,+0.0232]
& +175.5 [+161.2,+192.3]
& +286.9 [+277.7,+307.2]\\
ECI
& +0.0091 [-0.0005,+0.0233]
& +107.2 [+97.2,+126.2]
& +199.5 [+190.1,+220.1]\\
HopCPT
& -0.0244 [-0.0327,-0.0156]
& -8.9 [-30.2,+11.3]
& +48.2 [+31.0,+63.1]\\
\bottomrule
\end{tabular}}
\end{table*}

Because the Bike state explicitly contains a previous-week input, we
repeat the paired interval-score inference with blocks spanning 14
calendar days as a block-length sensitivity analysis. The 3859
admissible 14-day starting blocks contain 217--336 common targets, with
median 336. We retain $B=500$ bootstrap replications and the same random
seed. The point estimates are unchanged because only the bootstrap block
length is varied. Table~\ref{tab:app:bike-mbb14} compares the 7-day intervals with the 14-day sensitivity intervals. All seven comparator intervals remain strictly positive.

\begin{table*}[t]
\centering
\caption{Bike Sharing block-length sensitivity for paired interval-score
comparisons. Differences are comparator-minus-\SMSCP{} on the same 4169
targets. The point estimate $\Delta_{\mathrm{IS}}$ is unchanged across
block lengths; only the moving-block-bootstrap 95\% interval changes.}
\label{tab:app:bike-mbb14}
\scriptsize
\setlength{\tabcolsep}{5pt}
\begin{tabular}{lccc}
\toprule
Comparator
& $\Delta_{\mathrm{IS}}$
& 7-day 95\% CI
& 14-day 95\% CI\\
\midrule
Rolling & +315.4 & [+302.3,+344.7] & [+298.4,+348.1]\\
Recent-$k$ & +284.7 & [+274.6,+306.4] & [+274.2,+310.0]\\
Tuned-Recent & +296.2 & [+285.8,+312.8] & [+284.9,+313.9]\\
Tuned-NexCP & +276.3 & [+262.4,+304.5] & [+266.2,+307.5]\\
ACI-tuned & +286.9 & [+277.7,+307.2] & [+276.3,+309.4]\\
ECI & +199.5 & [+190.1,+220.1] & [+191.9,+220.6]\\
HopCPT & +48.2 & [+31.0,+63.1] & [+26.4,+63.8]\\
\bottomrule
\end{tabular}
\end{table*}

HopCPT also exhibits material initialization variability on this fixed
sequence. Across the 50 initializations, coverage has mean 0.870,
standard deviation 0.029, and range $[0.775,0.934]$; width has mean
317.5, standard deviation 63.1, and range $[235.8,519.4]$; interval
score has mean 467.7, median 443.0, standard deviation 85.3, and range
$[344.6,652.6]$. Sixteen initializations have lower interval score than
\SMSCP{} and 34 have higher interval score; 8 of 50 attain empirical
coverage at least 0.90. These summaries describe initialization
variability and are not time-series sampling confidence intervals.

\subsection{ResCP real-data comparison}
\label{app:rescp-realdata}

ResCP~\citep{neglia2026rescp} is evaluated separately on the real-data benchmarks
as a reservoir-state comparator using a matched-information-set adaptation of the upstream
reservoir and sampling procedure. The comparison uses the same frozen
point predictor, chronological outer candidate window, and evaluation
targets as the \SMSCP{} benchmark, while retaining ResCP's signed
residuals, reservoir-state similarity weighting and sampling, and
asymmetric interval construction. This is therefore an aligned
end-to-end comparison rather than a claim of reproducing the upstream
full forecasting pipeline exactly.

For Solar, profile selection is restricted to the three prespecified
upstream Solar configurations (\texttt{solar\_rnn},
\texttt{solar\_arima}, and \texttt{solar\_transformer}) and is performed
on the chronological validation block by mean interval score. Bike has
no upstream Bike-specific profile, so it uses only the upstream
\texttt{generic} configuration without dataset-specific profile search.
For each benchmark, target-level coverage indicators, widths, and
interval scores are averaged over three fixed ResCP random seeds
$(0,1,2)$ before paired time-series inference. The 7-day inference uses
the same moving-block-bootstrap construction with $B=500$ as the other
real-data comparisons.

\begin{table*}[t]
\centering
\caption{Real-data comparison with ResCP at nominal
$1-\alpha=0.90$. The last column reports
$\Delta_{\mathrm{IS}}=\mathrm{IS}_{\mathrm{ResCP}}
-\mathrm{IS}_{\mathrm{SMSCP}}$ with its pointwise paired 7-day
moving-block-bootstrap 95\% interval ($B=500$). Negative values indicate
lower interval score for ResCP. Coverage entries are empirical point
estimates and are interpreted relative to the nominal level. Width and
interval score are in the response units of the corresponding dataset
and should not be compared in magnitude across datasets.}
\label{tab:app:rescp-realdata}
\scriptsize
\setlength{\tabcolsep}{2.8pt}
\resizebox{\textwidth}{!}{%
\begin{tabular}{lrrr|rrr|c}
\toprule
& \multicolumn{3}{c|}{\SMSCP{}}
& \multicolumn{3}{c|}{ResCP}
& $\Delta_{\mathrm{IS}}$ [95\% CI]\\
Dataset & Cov. & Width & IS & Cov. & Width & IS & \\
\midrule
Alachua     & 0.910 & 264.0 & 361.0 & 0.886 & 202.4 & 334.0 & -27.0 [-31.9,-21.8]\\
Phoenix     & 0.920 & 188.2 & 252.8 & 0.883 & 90.8  & 188.5 & -64.4 [-69.9,-58.4]\\
Seattle     & 0.914 & 208.4 & 286.2 & 0.885 & 142.8 & 242.5 & -43.7 [-49.0,-38.9]\\
Houston     & 0.914 & 249.3 & 332.9 & 0.867 & 165.8 & 297.3 & -35.7 [-41.6,-30.3]\\
Minneapolis & 0.909 & 214.3 & 295.6 & 0.889 & 150.8 & 251.7 & -43.9 [-48.2,-39.3]\\
Bike        & 0.894 & 326.4 & 419.6 & 0.937 & 218.0 & 252.5 & -167.1 [-199.0,-146.3]\\
\bottomrule
\end{tabular}}
\end{table*}

Using $|\widehat{\mathrm{Cov}}-0.90|$ only as a descriptive
measure of proximity to the nominal level, the \SMSCP{} empirical
coverage point estimate is closer at Alachua, Seattle, Houston,
Minneapolis, and Bike, whereas ResCP is slightly closer at Phoenix.
This is not a formal test of coverage-calibration error, and higher
empirical coverage is not treated as inherently preferable. ResCP has
lower mean interval score on all six benchmarks, and all six paired
7-day moving-block-bootstrap intervals for
$\Delta_{\mathrm{IS}}$ are strictly negative.

For Bike, repeating the ResCP interval-score comparison with 14-day
blocks leaves the direction unchanged:
$\Delta_{\mathrm{IS}}=-167.1$ with 95\% interval
$[-207.5,-141.7]$, compared with $[-199.0,-146.3]$ under the 7-day blocks. This sensitivity analysis changes only the bootstrap block
length, not the point estimate.

\subsection{Computational diagnostics}
\label{app:runtime}

Table~\ref{tab:app:runtime} records execution diagnostics from the
same simulation runs. The timers do not define a controlled hardware
benchmark across all methods, so these values are not used to claim a
universal runtime ranking. Their role is only to document the
additional neural-training burden of the learned-retrieval comparator.

\begin{table*}[t]
\centering
\caption{Execution diagnostics from the simulation runs. HopCPT
times are per neural initialization; \SMSCP{} times are the median
wall time for the full \SMSCP{} path per data realization. These are
not controlled cross-method runtime benchmarks.}
\label{tab:app:runtime}
\scriptsize
\begin{tabular}{llrrr}
\toprule
Design & Size & \SMSCP{} median sec. & HopCPT median train+infer sec. & HopCPT train-time IQR\\
\midrule
HomoSin & S & 1.42 & 20.51 & 30.11 \\
HomoSin & L & 3.49 & 26.17 & 13.34 \\
DualFreq & S & 1.41 & 20.64 & 1.46 \\
DualFreq & L & 4.64 & 29.03 & 2.94 \\
HeteroSin & S & 1.06 & 18.55 & 4.42 \\
HeteroSin & L & 4.72 & 28.54 & 1.63 \\
ARCH(1) & S & 0.99 & 16.16 & 3.56 \\
ARCH(1) & L & 3.68 & 26.17 & 11.75 \\
MarkovSwitch & S & 0.97 & 15.49 & 9.46 \\
MarkovSwitch & L & 3.59 & 25.55 & 10.79 \\
\bottomrule
\end{tabular}
\end{table*}


\section{Data-Generating Processes}
\label{app:sim_dgp}

This appendix gives the complete specifications for the simulation
designs in the simulation panel. All designs use the same rolling
lag-one-correlation predictor based on the 20 most recent observed
responses and the score $R_t=|Y_t-\hat\mu_t|$. Core designs use both the small
split $(100,100,100,100)$ and large split $(300,300,300,300)$; the
additional simulation designs are evaluated only under the large split. All
non-neural cells use $B=500$ paired Monte Carlo replications.

Throughout this appendix, local-linear offsets are divided by
their empirical standard deviation within the one-sided history window,
with numerical floor $10^{-4}$. Ill-conditioned local least-squares
systems fall back to the local response mean with zero slope
coefficients. These are numerical implementation safeguards and do not
use the target response.

\subsection{Core simulation designs}

\paragraph{HomoSin.}
$Y_t=1.5\sin(t)+0.25\varepsilon_t$, with $\varepsilon_t\overset{\mathrm{iid}}{\sim}N(0,1)$.
The time grid is the unit-spaced integer grid $t=1,\ldots,T$.
The state input is chronological time, and
$Z_t=(\beta_0,\beta_1)$ is the one-sided local-linear level--slope
summary fitted on $s\in[t-h,t-1]$ using the centered time offset
$s-t$.

\paragraph{DualFreq.}
$Y_t=1.5\sin(t)+0.6\sin(\pi t/1.5)+0.2\varepsilon_t$, where $\varepsilon_t\overset{\mathrm{iid}}{\sim}N(0,1)$.
The state is the same strictly pre-target time-index level--slope state
as in HomoSin. The unit-spaced time parameterization is used at both
sample sizes.

\paragraph{HeteroSin.}
$Y_t=1.5\sin(t)+\sigma(t)\varepsilon_t$, where $\sigma(t)=0.25\{1+2.5\cos^2(t)\}$ and $\varepsilon_t\overset{\mathrm{iid}}{\sim}N(0,1)$.
The state is again the time-index level--slope state. Because the
response scale varies with phase while the state is constructed from
local response level and slope, the chosen time-index state captures the score-scale variation only
partially in this design.

\paragraph{ARCH(1).}
$Y_t=\varepsilon_t\sqrt{0.4+0.5Y_{t-1}^2}$, with $\varepsilon_t\overset{\mathrm{iid}}{\sim}N(0,1)$.
This is an ARCH(1), not a GARCH, process. The pre-target state input is
$Y_{t-1}^2$. For $s\in[t-h,t-1]$, the state
$Z_t=(\beta_0,\beta_1)$ is obtained from
\[
Y_s
=
\beta_0
+
\beta_1
\frac{Y_{s-1}^2-Y_{t-1}^2}{\hat\sigma_t}
+
\mathrm{error}.
\]
Although $Y_{t-1}^2$ contains the conditional-volatility information,
the state is constructed by locally regressing the raw response on this
input. Because $\mathbb E(Y_s\mid Y_{s-1}^2)=0$, the resulting
level--slope state need not preserve the volatility scale relevant to the
absolute prediction error. We therefore use ARCH(1) as a setting in which
the chosen state can miss information relevant to the target score
quantile.

\paragraph{MarkovSwitch.}
$Y_t=\delta S_t+\varepsilon_t$, with $\delta=2$ and $\varepsilon_t\overset{\mathrm{iid}}{\sim}N(0,1)$,
where $S_t\in\{0,1\}$ is a two-state Markov chain with
$\Pr(S_t=S_{t-1})=0.95$. The state input is $Y_{t-1}$ and
$Z_t=(\beta_0,\beta_1)$ is constructed by
\[
Y_s
=
\beta_0
+
\beta_1
\frac{Y_{s-1}-Y_{t-1}}{\hat\sigma_t}
+
\mathrm{error},
\qquad
s\in[t-h,t-1].
\]
This design creates abrupt transitions and serves as a boundary setting
for state matching based on smoothly recurring states.

\subsection{Additional simulation designs}

\paragraph{AR1.}
$Y_t=0.8Y_{t-1}+\varepsilon_t$, with $\varepsilon_t\overset{\mathrm{iid}}{\sim}N(0,1)$.
The state input is $Y_{t-1}$ and the level--slope state is constructed
from centered lagged-response offsets
$(Y_{s-1}-Y_{t-1})/\hat\sigma_t$ over the one-sided history window.

\paragraph{WindowAR.}
$Y_t=0.5Y_{t-5}-0.3Y_{t-6}+0.5\varepsilon_t$, with $\varepsilon_t\overset{\mathrm{iid}}{\sim}N(0,1)$.
The state input is $(Y_{t-5},Y_{t-6})$, and the three-dimensional
state $Z_t=(\beta_0,\beta_1,\beta_2)$ is fitted by
\[
Y_s
=
\beta_0
+
\beta_1\frac{Y_{s-5}-Y_{t-5}}{s_1}
+
\beta_2\frac{Y_{s-6}-Y_{t-6}}{s_2}
+
\mathrm{error},
\]
where $s_1$ and $s_2$ are the corresponding local empirical standard
deviations with floor $10^{-4}$.

\paragraph{CovARX.}
The exogenous inputs and response follow
\[
\begin{aligned}
X_{1,t}&=0.7X_{1,t-1}+0.4\sin(2\pi t/50)+\xi_{1,t},\\
X_{2,t}&=0.5X_{2,t-1}+0.3\cos(2\pi t/35)+\xi_{2,t},\\
Y_t&=-0.6+0.70Y_{t-1}-0.15Y_{t-2}
     +1.10X_{1,t}-0.60X_{2,t}+\varepsilon_t.
\end{aligned}
\]
Here $\xi_{1,t},\xi_{2,t}\sim N(0,0.25^2)$ and
$\varepsilon_t\sim N(0,0.20^2)$.
The observed pre-target inputs are $(X_{1,t},X_{2,t})$. The
three-dimensional state is fitted from
\[
\begin{aligned}
Y_s
&=
\beta_0
+
\beta_1\frac{X_{1,s}-X_{1,t}}{s_1}\\
&\quad+
\beta_2\frac{X_{2,s}-X_{2,t}}{s_2}
+
\mathrm{error},\\
&\hspace{2em}s\in[t-h,t-1].
\end{aligned}
\]
The local scaling and numerical floor are as specified above.

\end{document}